\documentclass{article}

\usepackage{microtype}
\usepackage{algorithm, algorithmic, amsmath}
\usepackage{multirow, multicol}
\usepackage{graphicx, color, float, subfigure, etoolbox, float, booktabs}

\usepackage{amsmath,amsfonts,amssymb,amscd,amsthm,bm,bbm}

\usepackage[dvipsnames]{xcolor}

\definecolor{darkblue}{rgb}{0.03, 0.27, 0.49}

\usepackage{hyperref}
\newtheorem{theorem}{Theorem}

\newtheorem{example}{Example}

\newtheorem{decisionrule}{Decision rule}

\newtheorem{remark}{Remark}

\newtheorem{postulate}{Postulate}
\newtheorem{assumption}{Assumption}

\def\equationautorefname~#1\null{Equation~(#1)\null}

\makeatletter
\newcommand\footnoteref[1]{\protected@xdef\@thefnmark{\ref{#1}}\@footnotemark}
\makeatother

\newcommand{\R}{{\mathbb R}}

\usepackage{enumitem}
\usepackage{titlesec}

\usepackage[accepted]{icml2020}

\icmltitlerunning{Bivariate Quantile Causal Discovery}

\begin{document}

\setlength{\parskip}{2pt}

\twocolumn[
\icmltitle{Distinguishing Cause from Effect Using Quantiles:  \\ Bivariate Quantile Causal Discovery}



\icmlsetsymbol{equal}{*}

\begin{icmlauthorlist}
\icmlauthor{Natasa Tagasovska}{unil,sdsc}
\icmlauthor{Val\'erie Chavez-Demoulin}{unil}
\icmlauthor{Thibault Vatter}{uc}
\end{icmlauthorlist}

\icmlaffiliation{unil}{HEC, University of Lausanne}
\icmlaffiliation{sdsc}{Swiss Data Science Center EPFL/ETHZ }
\icmlaffiliation{uc}{Statistics Department, Columbia University}

\icmlcorrespondingauthor{Natasa Tagasovska}{natasa.tagasovska@epfl.ch}

\icmlkeywords{Causality, Quantile Regression, Code Length, Minimum Description Length}

\vskip 0.3in
]



\printAffiliationsAndNotice{}  

\begin{abstract}
Causal inference using observational data is challenging, especially in the bivariate case.
Through the minimum description length principle, we link the postulate of independence between the generating mechanisms of the cause and of the effect given the cause to quantile regression.
Based on this theory, we develop Bivariate Quantile Causal Discovery (bQCD), a new method to distinguish cause from effect
assuming no confounding, selection bias or feedback.
Because it uses multiple quantile levels instead of the conditional mean only, bQCD is adaptive not only to additive, but also to multiplicative or even location-scale generating mechanisms.
To illustrate the effectiveness of our approach, we perform an extensive empirical comparison on both synthetic and real datasets.
This study shows that bQCD is robust across different implementations of the method (i.e., the quantile regression), computationally efficient, and compares favorably to state-of-the-art methods.

\end{abstract}

\section{Introduction}

Driven by the usefulness of causal inference in most scientific fields, an increasing body of research has contributed towards understanding the generative processes behind data. 
The aim is elevation of learning models towards more powerful interpretations: from correlations and dependencies towards \emph{causation} 
\cite{Pearl2000,Spirtes2000, dawid2007fundamentals, Pearl2009,scholkopf2019causality}.

While the golden standard for causal discovery is randomized control trials \cite{Fisher1936}, experiments or interventions in a system are often prohibitively expensive, unethical, or, in many cases, impossible. 
In this context, an alternative is to use observational data to infer causal relationships \cite{Spirtes2000, Maathuis2015}. 
This challenging task has been tackled by many, often relying on testing conditional independence and backed up by heuristics 
\cite{Maathuis2015,Spirtes2016,Peters2017}.

Borrowing from structural equations and graphical models, structural causal models \citep[SCMs, ][]{Pearl2009,Peters2017} represent the causal structure of variables $X_1, \cdots, X_d$ using equations such as 
\begin{align*}
X_c = f_c(X_{PA(c), \mathcal{G}}, N_c), \, c \in \left\{ 1, \dots, d\right\},
\end{align*}
where $f_c$ is a causal mechanism linking the child/effect $X_c$ to its parents/direct causes $X_{PA(c), \mathcal{G}}$,
$N_c$ is another variable independent of $X_{PA(c), \mathcal{G}}$,
and $\mathcal{G}$ is the directed graph obtained from drawing arrows from parents to their children.

Further complications arise when observing only two variables: one cannot distinguish between latent confounding ($X \leftarrow Z \rightarrow Y$) and direct causation ($X \rightarrow Y$ or $X \leftarrow Y$) without additional assumptions \cite{Shimizu2006, Janzing2012, Peters2014, Lopez-Paz2015}.
In this paper, we focus on distinguishing cause fom effect in a bivariate setting, assuming the presence of a causal link, but no confounding, selection bias or feedback.

An alternative is to impose specific structural restrictions.
For example, (non-)linear additive noise models with $Y = f(X) + N_Y$ \cite{Shimizu2006,Hoyer2009,Peters2011} 
or post nonlinear models $Y = g(f(X) + N_Y)$ \cite{Zhang2009} allow to establish causal identifiability without some of those assumptions.

\begin{figure}[t]
\centering
\vspace{-0.1cm}
\resizebox{0.45\textwidth}{!}{
 \includegraphics{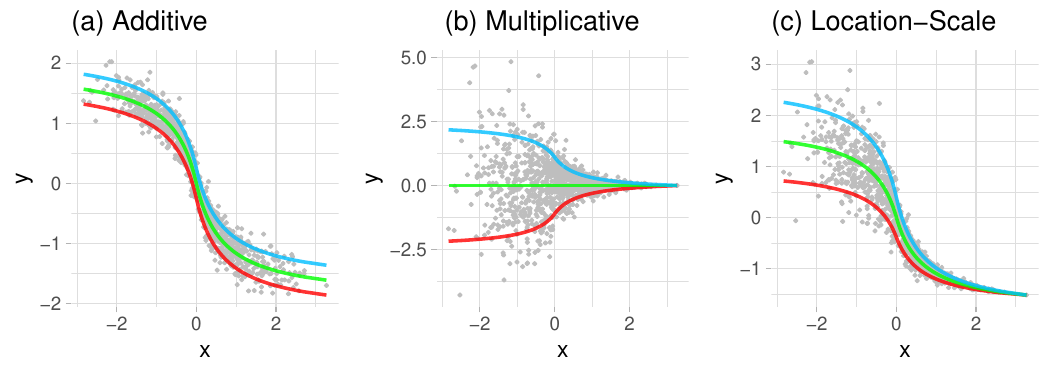}
 }
  \vspace{-0.5cm}
\caption{Diverse cause-effect ($X \rightarrow Y$) setups. Green - median, blue - $0.9^{th}$ quantile, red - $0.1^{th}$ quantile. Focusing on the mean only is not enough in the location-scale and multiplicative cases.}
\label{fig:mot_ex}
 \vspace{-0.5cm}
\end{figure}

In~\autoref{fig:mot_ex}, we illustrate a limitation of causal inference methods based on the conditional mean and refer to \autoref{sec:intuition} for more details on the underlying generative mechanisms.
From \autoref{fig:mot_ex}(a), it is clear that the variance of the effect variable is independent of the cause.
As a result, the independence between the cause and the effect's noise can be measured in various ways by substracting an estimate of the conditional mean.
In~\autoref{fig:mot_ex}(b) and (c) however, the mechanisms are $Y = g(X)N_Y$ and  $Y = f(X) + g(X)N_Y$ respectively.
In other words, the variance of the effect also depends on the cause, i.e. it exhibits heteroskedasticity.
Assume that $E(N_Y)=0$, then in the multiplicative case, we have that $E(Y|X=x)=0$.
And it is not sensible to use the conditional mean to identify the causal direction.
But since $\mbox{Var}(Y|X=x)=x^2$, the variance is more informative.
Similarly, features of the conditional distributions (e.g., conditional spread) different from the location might help.
In such cases, relying on mutliple (conditional) quantiles rather than on the mean only can help. 
As we show in~\autoref{sec:experiments}, it allows us to correctly determine the causal direction across various benchmarks where additive-noise competitors fail when their assumptions are not met.

\begin{figure}[t]
\centering
\resizebox{0.4\textwidth}{!}{
 \includegraphics{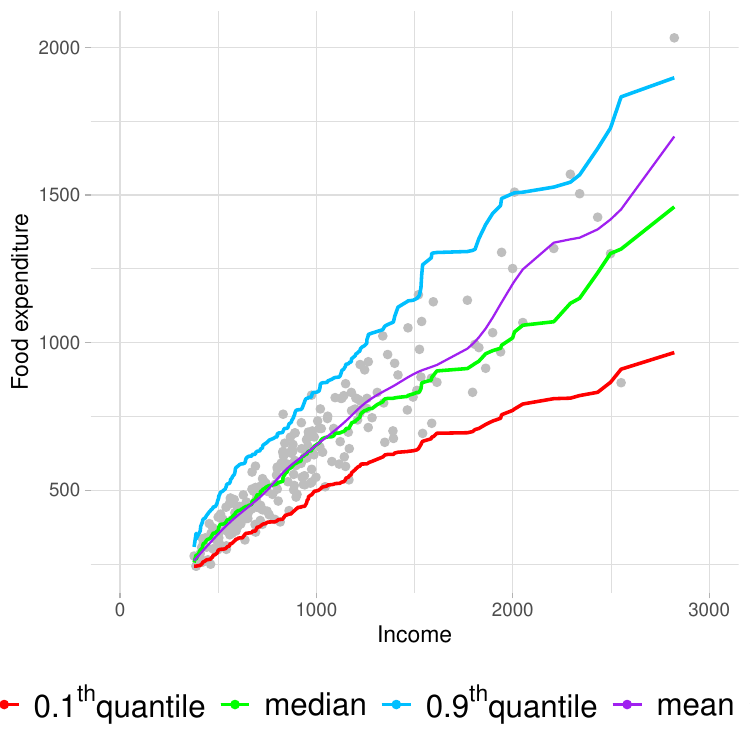}
 }
 \vspace{-0.5cm}
\caption{Heteroskedasticity in food expenditure as a function of income for Belgian working class households. Our method uses multiple quantiles to find the correct causal direction.}
\label{fig:engel_ex}
 \vspace{-0.5cm}
\end{figure}

A classic example of real world data displaying such features is the impact of income on expenditure on meals:
as an individual's income increases, so does its food expenditure and food expenditure variability.
An explanation could be as follows:
while a poorer person will spend a rather constant amount on inexpensive food, wealthier individuals occasionally buy inexpensive food and sometimes eat expensive meals.
Or, that wealthier individuals have more leeway when deciding which fraction of their income to allocate to food expenditure, whereas poorer ones are constrained by necessity.
This heteroskedastic effect can be observed in~\autoref{fig:engel_ex}, where our method use multiple quantiles to find the correct causal direction.

Another line of work avoids functional restrictions by relying on the 
\emph{independence of cause and mechanism} postulate \cite{Scholkopf2012,Peters2017}.
\begin{postulate}[\citealt{Sgouritsa2015}]
\label{postulate1}
The marginal distribution of the cause and the conditional distribution of the effect given the cause, corresponding to independent mechanisms of nature, are algorithmically independent (i.e., they contain no information about each other).
\end{postulate}
Information Geometric Causal Inference (IGCI) \cite{Janzing2012} uses the postulate directly for causal discovery.
Similarly, Regression Error based Causal Inference (RECI) \cite{blobaum2018cause} develops structural assumptions based on the postulate to imply that, if $X$ causes $Y$ implies an asymmetry in mean-regression errors, that is
\begin{align}\label{eq:MSEasym}
E[(Y-E[Y | X])^2] \leq E[(YX-E[X | Y])^2].
\end{align} 

Alternatively, \cite{Mooij2009,Janzing2010} reformulate the postulate through asymmetries in Kolmogorov complexities \cite{Kolmogorov1963} between marginal and conditionals distributions.
However, the halting problem \cite{Turing1938} implies that the Kolmogorov complexity is not computable, and approximations or proxies have to be derived to make the concept practical.

In this context, \cite{Mooij2010} proposes a method based on the minimum message length principle using Bayesian priors, while others are based on reproducing kernel Hilbert space embedding such as EMD \cite{Chen2014}, FT \cite{Liu2017} and KCDC by \citet{mitrovic2018causal}, or using entropy as a measure of complexity \cite{kocaoglu2017entropic}.
A related line of work suggests using the minimum description length \citep[MDL, ][]{Rissanen1978} principle as a proxy for Kolmogorov complexity: \cite{Budhathoki2017} uses MDL for causal discovery on binary data, and Slope \cite{Marx2017,  marx2019identifiability} implements (local and) global functional relations using MDL based regression and is suitable for continuous data. 

In this paper, we build on a similar idea, using \emph{quantile scoring} as a proxy for the Kolmogorov complexity through the MDL principle.
We develop a method leveraging asymetries similar to \eqref{eq:MSEasym}, replacing the squared loss and conditional mean by the pinball loss and conditional quantiles.
To the best of our knowledge, quantiles have only been mentioned in a somewhat related context by \cite{Heinze-Deml2017}, where quantile predictions are used to exploit the invariance of causal models across different environments.
As opposed to \cite{Heinze-Deml2017}, our method uses an asymmetry directly derived from the postulate, and therefore it does not require an additional variable for the environment.

To avoid the restrictive assumptions imposed by standard quantile regression techniques (e.g., linearity of the quantiles or additive relationships), 
we suggest fully nonparametric bQCD implementations using three different approaches: \emph{copulas, quantile forests or quantile neural networks}.
We also show that bQCD is robust to the choice regression approach.
To the best of our knowledge, we are the first to explore the idea of using conditional quantiles to distinguish cause from effect in bivariate observational data.
Our main contributions are:
\begin{itemize}
\item a new method based on quantile scoring to determine the causal direction without any assumptions on the class of causal mechanisms 
  along with a theoretical analysis justifying its usage (\autoref{sec:np_causal}),
\item \emph{quantile causal discovery} (bQCD), an efficient implementation robust to the choice of underlying regressor (\autoref{sec:bQCD_implementation}),
\item a new benchmark set of synthetic cause-effect pairs from additive, location-scale and multiplicative noise models (\autoref{sec:data}),
\item a comparative study to benchmark bQCD against state-of-the-art alternatives (\autoref{sec:experiments}).
\end{itemize}

\section{Causal Discovery using Quantiles}
\label{sec:np_causal}
In this section, we develop our quantile-based method for distinguishing between cause and effect from continuous and discrete observational data. 

\textbf{Problem setting}
We restrict ourselves to bivariate cases by considering pairs of univariate random variables.
We further simplify the problem by assuming the existence of a causal relationship but the absence of confounding, selection bias, and feedback.

\subsection{Kolmogorov Complexity and Quantile Scoring}
\label{sec:qs_to_kc}

Let $X$ and $Y$ be  two random variables with some joint distribution $F$, and where $F_X, F_Y$ and $F_{Y \mid X}, F_{X \mid Y}$ are respectively the marginal and conditional distributions.
The Kolmogorov complexity of a distribution $F$, denoted by $K(F)$, is the length of the shortest program $p$ that outputs $F(x)$ up to precision $q$ for all $x$, that is
\begin{align*}
 K(F)=\min \left\{ \mid p \mid \, :\, \mid U(p,q,x)-F(x) \mid \leq 1/q, \, \forall x \right\},
\end{align*}
with $U$ a Turing machine extracting $p(x,q)$ \citep[see][and references therein]{Janzing2010}.

In our setting, the Kolmogorov complexity can be leveraged to discover which one of $X$ and $Y$ is the cause and which one is the effect through the following theorem:
\begin{theorem}[\citealt{Mooij2010}]
\label{thm:kc_assym}
If~\autoref{postulate1} holds and $X$ causes $Y$, then $K(F_X) + K(F_{Y|X}) \leq K(F_Y) + K(F_{X|Y})$.
\end{theorem}

Stated differently, the causal direction between $X$ and $Y$ can be recognized as the least complex, that is the decomposition of the joint distribution leading to the lowest value of the Kolmogorov complexity.
This is because, in this context, the direct translation of~\autoref{postulate1} is that the mutual information between $F_X$ and $F_{Y|X}$ equals zero while that between $F_Y$ and $F_{X|Y}$ does not.
In other words, the number of bits saved when compressing $X$ and $Y$ jointly rather than compressing them independently is smaller when using the causal factorization of the joint distribution.
Importantly, because we assumed the existence of a causal link, the asymmetry in~\autoref{thm:kc_assym} is not only necessary but also sufficient.

\begin{remark}
Note that our problem setting corresponds to assumptions A, B and C in \cite{Mooij2010}.
Furthermore, \autoref{thm:kc_assym} holds up to an additive constant, as the asymmetry does not depend on the strings involved, but may depend on the Turing machines they refer to \citep[see e.g.,][]{Janzing2010, hernandez2010measuring}.
When using the same Turing machine to compute all complexities, carrying this constant over is not required.
\end{remark}

Since the Kolmogorov complexity is not computable, we use the MDL principle \citep{Rissanen1978} as a proxy: the less complex direction becomes the one allowing a better compression of the data.
In other words, the correct causal direction allows to store information using the shortest description length, or code length ($\mbox{CL}$).

\textbf{Two-step MDL encoding}
To construct a coding scheme satisfying the MDL principle, we use a two-stage approach \citep[see e.g.,][Section 3.1]{Hansen2001}: first encode a model and then the data using that model.
Assume that the goal is to compress $X = \left\{ X_i \right\}_{i=1}^n$ with $X_i\in \mathcal{X}$ i.i.d. according to some distribution $F_X \in \mathcal{F}$, where the model class is known to be $\mathcal{F} = \left\{ f_{\theta}( \cdot) \mid \theta \in \Theta \right\}$ and $\theta$ is an indexing parameter.
When $\theta$ is known, Shanon's source coding theorem implies that an encoding based on $f_{\theta}$ is ``optimal'': on average, it achieves the lower bound on any lossless compression, that is the differential entropy $-n \int_{\mathcal{X}} f_{\theta}(z) \log f_{\theta}(z) dz$. 
And the code length of a dataset encoded using the known $f_{\theta}$ is given by minus its log-likelihood $\mbox{CL}(X \mid \theta) = -\sum_{i=1}^n \log f_{\theta}(X_i)$.
This optimal scheme thus amounts at transmitting the true parameter along with the data encoded using its known distribution:
\begin{align}\label{eq:cl_two_part}
 \mbox{CL}_{\theta}(X) =  \mbox{CL}(\theta) + \mbox{CL}(X \mid \theta),
\end{align}
where the two terms on the right-hand side result from transmitting the model and the data encoded using the model \citep[see e.g.,][Section 3.1]{Hansen2001}.

\textbf{Encoding marginals via quantiles} Since knowledge of the model class and true parameter is seldom achieved, consider compression using partial information about the data-generating process.
Assuming that we only know a $\tau$-quantile $q_{X,\tau}= \operatorname{arg\,inf}_{q} \left\{q \mid F_X(q) = \tau \right\}$, we can compress the data by transmitting $\tau$, $q_{X,\tau}$, and the ``residuals'' $E_X = \{ X_i - q_{X,\tau} \}_{i=1}^n$.
This encoding results in
\begin{align} \label{eq:cl_tau_x}
  \mbox{CL}_{\tau}(X)&= \mbox{CL}(\tau) + \mbox{CL}(q_{X,\tau}) + \mbox{CL}(E_X \mid q_{X,\tau},\tau ).
\end{align} 
To encode $E_X$, it is then natural to use the asymmetric Laplace (AL) distribution \citep[see][]{Aue2014,geraci2006,yu2003} with density $f(z; q, \tau) = \tau ( 1- \tau) \exp (-S_{\tau}(q, z))$, where $S_{\tau}(\cdot, \cdot)$ is the quantile scoring (QS) function 
\begin{align*}
  S_{\tau}(x_1, x_2) = (\mathbb{I}\left\{x_1 \geq x_2 \right\} - \tau)(x_1 - x_2).
\end{align*}
Given that the distribution of $E_X$ is generally unknown, using the AL is optimal in the sense that the population quantile is exactly the minimizer of the QS's expected value, namely $q_{X,\tau} =\mbox{argmin}_{q}\, \mathbb{E}\left[S_{\tau}(q, X) \right]$.
We revisit the link between QS and the AL in~\autoref{sec:intuition}.

Using this encoding, the last term in the right-hand side of~\eqref{eq:cl_tau_x} is finally given by the negative of the AL log-likelihood \citep{Rissanen1986}, that is
\begin{align} \label{eq:cl_e_x}
 \mbox{CL}(E_X \mid q_{X,\tau},\tau ) = \sum_{i=1}^n S_\tau(q_{X,\tau}, X_i) - a_n(\tau),
\end{align}
where $a_n(\tau)=n\log(\tau (1-\tau))$.

\textbf{Encoding conditionals via quantiles} Next, consider the problem of compressing $\left\{ X_i, Y_i \right\}_{i=1}^n$ with $(X_i, Y_i)$ i.i.d. according to some joint distribution $F$.
Because $F$ can be decomposed using the marginal and the conditional distributions, one can proceed as above with $F_X$ to compress $X$, and then use  $F_{Y \mid X}$ to compress $Y$ given $X$.
Assume similarly that the only information about the conditional data-generating process is the conditional $\tau$-quantile $q_{Y \mid X = x,\tau} = \operatorname{arg\,inf}_{q} \left\{q \mid F_{Y \mid X = x}(q) = \tau \right\}$ of $Y$ given $X = x$.
Encoding a conditional distribution using~\eqref{eq:cl_tau_x} and~\eqref{eq:cl_e_x} thus results in 
\begin{align} \label{eq:cl_tau_y_x}
 \mbox{CL}_{\tau}(Y \mid X)&= \mbox{CL}(\tau) + \mbox{CL}(q_{Y \mid X,\tau}) \\
 &\phantom{=}+ \sum_{i=1}^n S_\tau(q_{Y \mid X = X_i,\tau}, Y_i) - a_n(\tau). \nonumber
\end{align}
In other words, the data is compressed by transmitting $\tau$, $q_{Y \mid X,\tau}$, and the ``residuals'' $E_{Y \mid X} = \{ Y_i - q_{Y \mid X = X_i,\tau} \}_{i=1}^n$.
Note that, for a given fixed quantile level $\tau$, while $q_{X,\tau}$ is a real number, $q_{Y \mid X,\tau}$ is generally a function of the conditioning variable.

\textbf{Causal identification via quantiles}
The same idea can be applied to compress the data with the decomposition of the joint distribution $F$ using the marginal $F_Y$ and the conditional $F_{X \mid Y}$ distributions.
But according to~\autoref{thm:kc_assym} and using $\mbox{CL}$s as proxies for Kolmogorov complexities, if $X$ is a cause of $Y$, one expects that for all $\tau \in (0,1)$
\begin{align*}
  \mbox{CL}_{\tau}(X) +  \mbox{CL}_{\tau}(Y \mid X) \leq  \mbox{CL}_{\tau}(Y) +  \mbox{CL}_{\tau}(X \mid Y)
\end{align*}
with high probability as the sample size increases.
We thus make the following ``identifying'' assumption for $X$ to be a cause of $Y$:
\begin{assumption}
$P \left( \frac{\mbox{CL}_{\tau}(X) +  \mbox{CL}_{\tau}(Y \mid X) }  {\mbox{CL}_{\tau}(Y) +  \mbox{CL}_{\tau}(X \mid Y)} \leq 1 \right) \underset{n \to \infty}{\longrightarrow} 1$
\label{assumption1}
\end{assumption}
This assumption is called identifying because, if the ratio of CLs equals 1 for all $\tau \in \left( 0,1 \right)$, quantile-based CLs cannot be leveraged for causal discovery.
For any given sample size and quantile level $\tau$, cases where $\mbox{CL}(\tau) = \infty$, $\mbox{CL}(q_{X,\tau})  = \infty$ or $\mbox{CL}(q_{Y \mid X,\tau})  = \infty$ are problematic.
The issue can be resolved by assuming for instance that all population quantities are computable and can be transmitted using a finite albeit potentially increasing precision.

Using $O$/$o$ for the usual asymptotic notations, we draw a first link between CLs and QSs at the population level.
\begin{theorem}\label{thm:cl_to_qs}
Assume that $\mbox{CL}(l)=o(n)$ for $l \in \{\tau,  q_{X, \tau}, q_{Y, \tau}, q_{Y \vert X, \tau}, q_{X \vert Y, \tau}) \}$, then~\autoref{assumption1} holds
if and only if
\begin{align}\label{eq:identification_inequality}
\frac{\mathbb{E} \left[ S_\tau(q_{X,\tau}, X_i) \right] + \mathbb{E} \left[ S_\tau(q_{Y \mid X = X_i,\tau}, Y_i) \right]}{\mathbb{E} \left[ S_\tau(q_{Y,\tau}, Y_i) \right]+ \mathbb{E} \left[ S_\tau(q_{X \mid Y = Y_i,\tau}, X_i) \right]}\leq 1 .
\end{align}
\end{theorem}
Because the population (conditional) quantiles and the expectations involved in the ratio are seldom known in practice, \autoref{thm:cl_to_qs} cannot be leveraged directly for causal discovery.
However, it can be used to determine whether a given causal model satisfies~\autoref{assumption1}, as exmplified below.
The proof can be found in \autoref{sec:proof_thm2}.

\begin{example}
Consider $X$ being a cause of $Y$ in the linear model defined by $X = \theta_1 N_X$ and $Y = \gamma X + \theta_2 N_Y$ with $N_X$ and $N_Y$ two independent sources of noise and $\theta_1, \theta_2, \gamma > 0$ such that $\mbox{var}(X) = \mbox{var}(Y)$.
Note that the condition on the variances simply ensures that the variables have the same scale.

Letting $N_X, N_Y \sim N(0,1)$, $X \sim N(0, \theta_1^2)$ and $Y \sim N(0, \gamma^2\theta_1^2 + \theta_2^2)$. 
The variance equality condition can be satisfied by choosing $\gamma \in (0,1)$ and letting $\theta_2 = \theta_1 \sqrt{1 - \gamma^2}$, resulting in
$X\mid Y = y \sim N( \gamma y, (1 - \gamma^2) \theta_1^2)$ and $Y\mid X = x \sim N( \gamma x, (1 - \gamma^2) \theta_1^2)$.
Using the fact that, if $Z \sim N(\mu, \sigma)$, then $\mathbb{E} \left[ S_\tau(q_{Z,\tau}, Z) \right] = \sigma c(\tau)$,
where $c(\tau) = e^{-\Phi^{-1}(\tau)^2/2}/\sqrt{2 \pi}$ with $\Phi$ the standard normal cumulative distribution,
it is then straightforward to verify that the ratio of expectations in \autoref{thm:cl_to_qs} is equal to one independently of $\tau$.
In other words, the linear Gaussian model is not identifiable \citep{Shimizu2006, Hoyer2009, Peters2014}.
\end{example}

\begin{remark}\label{remark:structural}
Developing structural or distributional assumptions, e.g. based directly on \autoref{postulate1}, such that \autoref{assumption1} or \eqref{eq:identification_inequality} holds is an open problem.
\cite{blobaum2018cause} paves the way in the context of \eqref{eq:MSEasym} and in the regime of almost deterministic relations.
But quantiles and quantile scores are harder to manipulate than conditional variances and squared losses.
\end{remark}

\textbf{Causal discovery via quantiles}
In order to leverage~\autoref{assumption1} for causal discovery, we let $\widehat{q}_{X, \tau}$, $\widehat{q}_{Y, \tau}$, $\widehat{q}_{X \vert Y, \tau}$, $\widehat{q}_{Y \vert X, \tau}$ be estimators of the respective population quantiles, and further make the following assumption:
\begin{assumption} $\widehat{q}_{X, \tau}$, $\widehat{q}_{Y, \tau}$, $\widehat{q}_{X \vert Y, \tau}$, $\widehat{q}_{Y \vert X, \tau}$ satisfy
\begin{itemize}
\item $\left| \widehat{q}_{X, \tau} - q_{X, \tau} \right| = o_p(1)$ and $\left| \widehat{q}_{Y, \tau} - q_{Y, \tau} \right| = o_p(1)$,
\item $\left| \widehat{q}_{Y \mid X = x, \tau} - q_{Y \mid X = x, \tau} \right| = o_p(1)$ for every $x$ and\\ $\left| \widehat{q}_{X \mid Y = y, \tau} - q_{X \mid Y = y, \tau} \right| = o_p(1)$ for every $y$,
\item  $\mbox{CL}(\widehat{q}_{X, \tau}) = o(n)$, $\mbox{CL}(\widehat{q}_{Y \vert X, \tau}) = o(n)$, \\$\mbox{CL}(\widehat{q}_{Y, \tau}) = o(n)$, and $\mbox{CL}(\widehat{q}_{X \vert Y, \tau}) = o(n)$,\vspace{0.15cm}
\end{itemize}
using $O_p$/$o_p$ for stochastic boundedness and convergence in probability. 
  \label{assumption2}
\end{assumption}

The first two bullet points simply state that the unconditional and conditional quantile estimators are consistent without rate.
As for the third bullet point, note that having the same growth rate for the CL of all models does not prevent a smaller number of parameters in the correct causal direction \citep[see e.g.,][]{blobaum2018cause, marx2019identifiability}.
However, it means that, if the population quantiles $q_{\cdot, \tau}$ are replaced by estimators $\widehat{q}_{\cdot, \tau}$ in \eqref{eq:cl_tau_x} and \eqref{eq:cl_tau_y_x}, the CLs of the models are all asymptotically dominated by the CLs of the residuals, namely the $\sum_{i=1}^n S_\tau(\cdot, \cdot)$ terms.

The third bullet point of \autoref{assumption2} includes the important case where all CLs are $O(\log n)$ \citep{Rissanen1983,Rissanen1986}:
discretizing a compact parameter space with a $n^{-1/2}$ grid (i.e., the magnitude of the estimation error) and transmitting an estimated parameter using a uniform encoder with this precision is optimal for regular parametric families.
Using this $n^{-1/2}$ precision, each parameter thus leads to a cost of $1/2\,\log n$.

For nonparametric (i.e., with a non-Euclidean parameter space) models, a similar idea can be applied, as estimators typically converge at a rate slower than $n^{-1/2}$.
For instance, it is well known that the optimal grid size for histogram estimators of continuous densities with bounded first derivative is proportional to $n^{-1/3}$ \citep[Theorem 6.11,][]{wasserman2006all}.
And if each of the histogram heights is encoded using $n^{-1/2}$ grid (i.e., smaller than the estimation error), the resulting estimator's CL is proportional to $n^{1/3}/2\,\log n$, which satisfies the third bullet point of \autoref{assumption2}.
More generally, the fastest possible rate for kernel density estimators of continuous densities with $k$ bounded derivatives is $n^{k/(2k+1)}$ \citep[Theorem 6.31,][]{wasserman2006all}. 
As a result, discretizing the support on a grid proportional to $n^{-k/(2k+1)}$ and encoding the kernel values using a $n^{-1/2}$ grid encures an estimator's CL proportional to $n^{k/(2k+1)}/2\,\log n$.

However, \autoref{assumption2} excludes datasets where quantiles can only be consistently estimated with models having ``too many parameters''.
For instance, if the only consistent estimator is an over-parametrized neural network with $o(n_{layers} \times n_{neurons/layer})>o(n)$, then \autoref{assumption2} does not hold.
But such a situation is unlikely in our bivariate setting.

We can then state the following theorem: 
\begin{theorem}
\label{thm:qs_assym}
Under~\autoref{assumption2},~\autoref{assumption1} holds
if and only if \vspace{-0.2cm}
\begin{align*} 
P\left( \frac{S_{X, \tau}  +  S_{Y \mid X, \tau}}{ S_{Y, \tau}  +  S_{X \mid Y, \tau} }  \leq 1 \right) \underset{n \to \infty}{\longrightarrow} 1,
\end{align*}
with the scores $S_{X, \tau} = \sum_{i=1}^n S_\tau(\widehat{q}_{X,\tau}, X_i)$, $S_{Y \mid X, \tau} = \sum_{i=1}^n S_\tau(\widehat{q}_{Y \mid X = X_i,\tau}, Y_i)$ and similarly for $Y$ and $X \mid Y$.
\end{theorem}

Note that~\autoref{thm:qs_assym} does not state that CLs and QSs are equivalent, but rather that inequalities in CLs imply inequalities in QSs with respect to a specific statistical model, and conversely. 
In other words,~\autoref{thm:qs_assym} implies that there is an equivalence between minimizing code length and quantile score. 
Hence, because of the MDL principle, the causal direction can be inferred from the lowest quantile score. 
The proof can be found in the supplementary material (\autoref{sec:proof_thm3}), but the intuition is as follows.

Because of the third bullet point in~\autoref{assumption2}, QSs, namely the CLs of the residuals, asymptotically dominate the CLs of the models.
As a result, using QSs corresponding to consistent models is sufficient for causal discovery.

Thanks to the stability (or invariance) of the true causal model, we expect \autoref{assumption1} to hold over different quantile levels.
However, since a single quantile is generally not enough to characterize a distribution, we further consider
\begin{align}
\widehat{S}_{X} = \int_{[0,1]}  S_{X, \tau} d\tau,
\end{align}
\label{eq:score}
and similarly for $X \mid Y$, $Y$ and $X \vert Y$.
By pooling results at different quantile levels, we aim at better describing the marginal and conditional distributions.
Arguing that estimating high and low (conditional) quantiles is hard, we could use only quantiles close to the median, that is integrating between over $[0.4,0.6]$ instead of $[0,1]$). 
We empirically found that this was more error prone when the generative models have asymmetries or multiplicative noises.
Finally, we use averaging through integration rather than the maximal QS difference over quantile levels because the scale of QS is not uniform (e.g., the closer to 0.5 the higher).
Hence, using maximization would essentially mean basing the decision on the median only, whereas properly capturing the spread of the data is also important.

\begin{decisionrule}[Bivariate Quantile Causal Discovery]
\label{thm:causal_rule}
Let $S_{X \to Y} = \widehat{S}_{X}+\widehat{S}_{Y \vert X}$ and $S_{Y \to X} = \widehat{S}_{Y}+\widehat{S}_{X \vert Y}$
If $S_{X \to Y} < S_{Y \to X}$, conclude that $X$ causes $Y$. If $S_{X \to Y} > S_{Y \to X}$, conclude that $Y$ causes $X$. Otherwise, do not decide.
\end{decisionrule}

\subsection{Intuition}\label{sec:intuition}

The mean and squared loss duality leads to using the MSE as a metric for mean regression even for non-Gaussian data.
And the pinball loss plays a similar role in quantile regression.
In the MDL paradigm, the optimal encoding uses the data's true distribution, which is usually unkown for the residuals of either a mean or quantile regression.
Without knowing this optimal encoding, it is sensible to use a distribution related to the loss that is minimized.
A Gaussian or asymmetric Laplace (AL) encoding should thus not be seen as optimal, but as a ``default'' related to the specific mean or quantile regression problem.
Furthermore, a Gaussian encoding is unlikely to be appropriate for quantile residuals, which are often asymmetrically distributed.

To make this idea more precise, consider \eqref{eq:MSEasym}, albeit from the point of view of the MDL principle.
One seeks a model that allows the best compression of the data, measured in code length (CL).
Using a two-part scheme as in \eqref{eq:cl_two_part}, the total CL is the sum of the conditional mean's CL and that of the residuals, which is equal to the negative log-likelihood of the distribution used to encode them.
And encoding the residuals using the normal distribution is natural in the same sense as using the mean squared error as a regression metric.
The reason is that minimizing the squared loss is equivalent to maximizing the Gaussian log-likelihood, thus minimizing the residuals' CL. 

A similar reasoning can be applied to quantile regression, where encoding the residuals using an AL distribution \cite{Koenker1999} represents a sensible default. 
This is because minimizing the quantile score (QS) is equivalent to maximizing the AL log-likelihood ($a_n(\tau) - \mbox{QS}$), thus minimizing the residuals' CL.
Intuitively, the likelihood corresponding to (conditional) residuals in the causal direction is higher, that is the QS and CL are smaller: the shortest CL corresponds to the largest AL likelihood/smallest QS, which establishes a link between minimizing QS and the MDL principle.

\begin{figure}
\centering
\resizebox{0.45\textwidth}{!}{
 \includegraphics{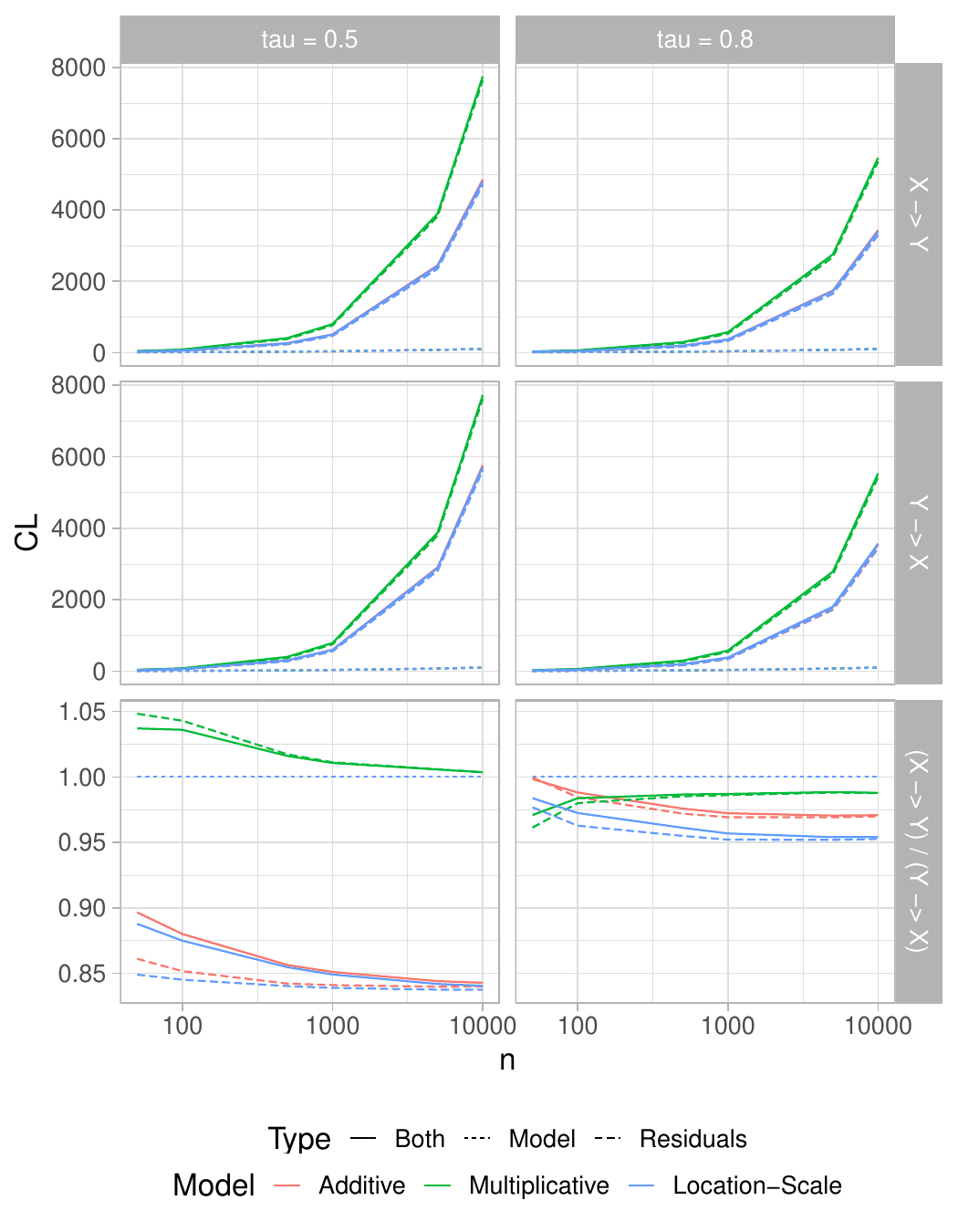}
 }
 \vspace{-0.5cm}
\caption{Verifying \autoref{assumption1} for the setups of \autoref{fig:mot_ex}. The CLs of the residuals asymptotically dominate the CLs of the marginal and conditional models. The multiplicative model is not identifiable using the median only. The correct causal direction has a lower CL.}
 \vspace{-0.5cm}
\label{fig:assumption_ex}
\end{figure}

Let us now illustrate the key ideas in the theoretical justification of bQCD.
In \autoref{fig:assumption_ex}, we revisit the toy examples from \autoref{fig:mot_ex}, namely, the different setups: additive, multiplicative and location-scale causal pairs.
We disentangle the CL's different components for both the causal $X \rightarrow Y$ and anti-causal $Y \rightarrow X$  direction, for two quantile levels $\tau \in \lbrace 0.5, 0.8 \rbrace$ and per generative model.

Curves are obtained by averaging over 100 repetitions.

As described in~\autoref{sec:qs_to_kc}, causal discovery by MDL involves four summands for each possible direction.
For instance, for $X \to Y$, we have
\begin{itemize}
\item two for the marginal and conditional model parameters, $\mbox{CL}(\widehat{q}_{X,\tau})$ and $ \mbox{CL}_{\tau}(\widehat{q}_{Y \mid X,\tau})$,
\item and two for the marginal and conditional residuals, $\mbox{CL}(E_X \mid \widehat{q}_{X,\tau},\tau )$ and $\mbox{CL}(E_{X | Y}  \mid \widehat{q}_{X | Y, \tau},\tau)$.
\end{itemize}
\vspace{0.3em}
In \autoref{fig:assumption_ex}, we use \emph{Model} for the former and \emph{Residuals} for latter and sum them by type.
The sum of both model and residuals CLs is given by full lines.
Unconditional and conditional quantiles are estimated respectively using the empirical quantile and  kernel quantile regression.
A uniform encoding is assumed in both cases, resulting in respective model CLs of $1/2 \log(n)$ and $n^{1/3}/2 \log(n)$.

In the top four panels of \autoref{fig:assumption_ex}, it is clear that the CLs of the residuals (i.e., the QSs) dominate the CLs of the marginal and conditional models.
This illustrates the third bullet point in \autoref{assumption2}.

In the bottom two panels of \autoref{fig:assumption_ex}, we see that the ratio of causal to anti-causal CLs generally\footnote{Because we assumed the same uniform encoding for both directions, the ratio of CLs of models is always equal to one.} converges to some constant smaller than 1, 
which illustrates \autoref{assumption1}.
It also means that \autoref{thm:causal_rule} would lead to the correct causal direction.
The multiplicative noise mechanism has an interesting behaviour:
considering the median only (i.e., $\tau = 0.5$), the ratio converges slowly to one, and therefore, the causal direction is not identifiable. However given that the ratio converges to some constant smaller than one for $\tau=0.8$, pooling the decision over multiple quantiles as in \autoref{thm:causal_rule} would again lead to the correct causal direction.

\section{Experiments}
\label{sec:experiments}

\subsection{bQCD implementation}
\label{sec:bQCD_implementation}

\textbf{Quantile regression}
\autoref{thm:qs_assym} holds provided that the model is consistent and its complexity does not grow too fast.
As such,~\autoref{thm:causal_rule} can be implemented using any quantile regression approach that satisfies \autoref{assumption2}.
In our experiments, we use three methods, namely nonparametric copulas \citep{Geenens2017}, quantile forests \cite{meinshausen2006quantile}, quantile neural networks \cite{Cannon2018}, and show that they yield qualitatively and quantitatively similar results.
We refer to \autoref{sec:quant_copula_details} for more details on the regression methods and their specific implementations.

For estimating the overall quantile scores (aggregating multiple quantile levels), we use Legendre quadrature to approximate the 
integral over $[0,1]$, as it is fast and precise for univariate functions.
In other words, denoting by $\left\{ w_j, \tau_j\right\}_{j=1}^m$ the $m$ pairs 
of quadrature weights and nodes, we use $\int_0^1 g(\tau) d\tau \approx 
\sum_{j=1}^m w_j g(\tau_j)$, which when plugged into \eqref{eq:score} yields $\widehat{S}_{X} = \sum_{j=1}^m w_j \widehat{S}_{X}(\tau_j)$, $\widehat{S}_{Y} = \sum_{j=1}^m w_j \widehat{S}_{Y}(\tau_j)$, $\widehat{S}_{X \vert Y} = \sum_{j=1}^m w_j \widehat{S}_{X \vert Y}(\tau_j)$, and $\widehat{S}_{Y \vert X} = \sum_{j=1}^m w_j \widehat{S}_{Y \vert X}(\tau_j)$.
Summing over an equally spaced grid with uniform weights or using quadrature nodes and weights yields two valid approximations of an integral, and using one or the other should not matter. 
But the quadrature gives more importance to the center of the distribution (i.e., quantiles closer to 0.5 have a higher weight).
Note that, to compute scores free of scale bias, the variables are transformed to the standard normal scale.

\textbf{Computational complexity} 
bQCD scales linearly with the size of input data,  $O(n)$  for copulas, $O(n_{tree} \times n_{attr} \times depth \times   n)$ for random forests and $O(epochs \times n_{weights}  \times n)$ corresponding to the choice of regressor. 
However, using multiple quantile levels does not increase the complexity significantly since all of the suggested implementations allow for simultaneous conditional quantile estimation, that is, we estimate a single model at each possible causal direction~\footnote{Except for bQCD, where the nature of copula models allows for joint estimation.}, from which we compute all of the requested quantiles.
Roughly speaking, since $n >> m$, the overall complexity scales with $\mathcal{O} (n)$.
As such, bQCD compares favorably to nonparametric methods relying on computationally intensive procedures, for an instance based on kernels  \cite{Chen2014, Hernandez-Lobato2016a} or Gaussian processes \cite{Hoyer2009, Mooij2010, Sgouritsa2015}.

The parameter $m$ can be used to control for the trade-off between the computational complexity and the precision of the estimation.
We recommend the value $m = 3$ which, makes it possible to capture variability in both location and scale.
Setting $m = 1$ is essentially equivalent to using only the conditional median for causal discovery, a setting that suitable for distributions with constant variance.
An empirical analysis of the choice of $m$ is provided in the following section.
In what follows, we report results for bQCD with $m = 3$ if not stated otherwise.

\subsection{Datasets, baselines and metrics}
\label{sec:data}
\textbf{Benchmarks}
For simulated data, we first rely on the following scenarios \cite{Mooij2016}: \emph{SIM} (without confounder), \emph{SIM-ln} (with low noise), \emph{SIM-G}  (with distributions close to Gaussian), and \emph{SIM-c} (with latent confounder).
There are 100 pairs of size $n = 1000$ in each of these datasets. 

As a second benchmark, inspired by \cite{Peters2014}, we generate a diverse dataset of additive, location-scale and multiplicative causal pairs.
We include nonlinear additive noise (\emph{AN}) models of the form $Y = f(X) + E_Y$ for some deterministic function $f$ with $E_Y \sim {\cal{N}}(0,\sigma)$, $X \sim {\cal{N}}(0,\sqrt{2})$, and $\sigma \sim {\cal{U}}[1/5, \sqrt{2/5}]$.
In \emph{AN}, $f$ is an arbitrary nonlinear function simulated using Gaussian processes  \citep[GP,][]{rasmussen2006gaussian} with a Gaussian kernel of bandwidth one.
Since the functions in \emph{AN} are often non-injective, we include \emph{AN-s} to explore the behavior of bQCD in injective cases.  
In this setup, $f$ are sigmoids as in \cite{Buhlmann2014}.
The third experiment considers location-scale (\emph{LS}) data generating processes with both the mean and variance of the effect being functions of the cause, that is $Y = f(X) + g(X) E_Y$, and $E_Y$ and $X$ are similar as for the additive noise models. 
\emph{LS} and \emph{LS-s} then correspond to the Gaussian processes and sigmoids described for \emph{AN} and \emph{AN-s}.
Finally, the fourth experiment considers multiplicative models (\emph{MN}) as $Y = f(X) E_Y$, with $f(X)$ sampled as sigmoid functions and $E_Y \sim {\cal{U}}(0,1)$.
In each of the second, third, and fourth experiments, we simulate 100 pairs of size $n = 1000$.
All pairs have equal weights with variable ordering according to a coin flip, therefore resulting in balanced datasets.
Example datasets for each of the simulated experiments are shown in the supplementary material (\autoref{sec:simulated_pairs}).

For real data, we use the \href{webdav.tuebingen.mpg.de/cause-effect/}{T\"{u}bingen 
CE benchmark (version Dec 2017)}, consisting of 108 pairs from 37 different domains, from which we consider only the 99 pairs that have univariate continuous or discrete cause and effect variables. 
When evaluating the performance on this dataset we included the pairs' corresponding weights which accounts for potential bias in cases where pairs were selected from same multivariable dataset.

\textbf{Baselines}
On simulated data, we compare bQCD to state-of the-art approaches, namely RESIT \cite{Peters2014}, biCAM \cite{Buhlmann2014}, LinGaM \cite{Shimizu2006}, and GR-AN \cite{Hernandez-Lobato2016a}, which are ANM-based, and IGCI \cite{Janzing2010}, EMD \cite{Chen2014}, RECI \citep{blobaum2018cause}, Slope \cite{Marx2017} and Sloppy \citep{marx2019identifiability}, based on the independence postulate.
We also consider other methods such as PNL-MLP \cite{Zhang2009}, GPI \cite{Mooij2010}, ANM \cite{Hoyer2009}, and CURE \cite{Sgouritsa2015}.
All baselines distinguish between cause and effect solely in the bivariate case, except for CAM, RESIT and LinGaM whose impementations allow for higher dimensional causal discovery.
Implementation details and hyper parameters for all baselines are described in the supplementary material (\autoref{sec:impl_details}). 

Our code and datasets are available in the submitted supplementary package and {\color{blue} https://github.com/tagas/bQCD}.

\textbf{Evaluation metrics}
As \cite{Mooij2016}, we use the \emph{accuracy} for \emph{forced decisions} and the \emph{area under the receiver operating curve} (ROC) for \emph{ranked decisions}.  
The former corresponds to forcing the compared methods to decide the causal direction. The later corresponds to using heuristic scores allowing to rank confidence in the directions along with ROC/AUC as performance measure.
As a confidence heuristic for the ranked decisions for bQCD, we use same score as (24) in \cite{Mooij2016}, that is $\hat{C} = -S_{X \to Y} + S_{Y \to X}$, with higher absolute values corresponding to higher confidence.
This score can also be paired with the null-hypercompressibility inequality \citep{grunwald2007minimum} to derive a significance test and abstain from making insignificant inferences.

\subsection{Results and discussion}

\begin{figure}
\centering
\resizebox{0.3\textwidth}{!}{
 \includegraphics{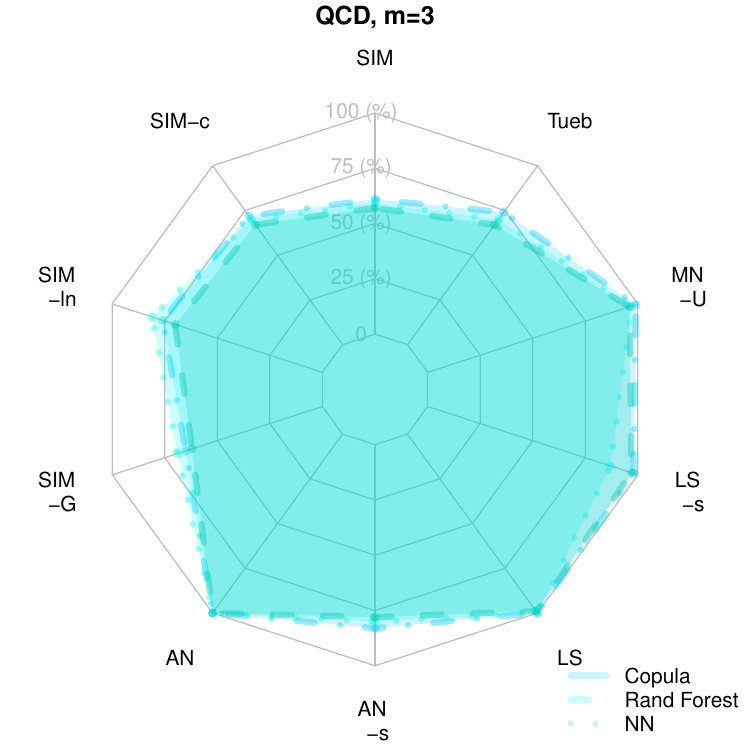}
 }
 \vspace{-0.5cm}
\caption{bQCD achieves consistent results across all benchmarks regardless of the practical implementation. }
 \vspace{-0.5cm}
\label{fig:bQCD_m3}
\end{figure}

\textbf{bQCD robustness wrt to its implementation} 
From~\autoref{fig:bQCD_m3} it is clear that the choice of implementation in estimating the conditional quantiles has no significant impact and bQCD provides consistent results across all benchmarks.
Further results that confirm the robustness to implementation for different values of $m$ are shown in \autoref{sec:quant_copula_details}, \autoref{fig:bQCD_all_m}. 
In the remaining of the paper we will show the copula-based results.

\begin{figure}[h!]
 \vspace{-0.5cm}
\resizebox{0.5\textwidth}{!}{
\centering
 \includegraphics[width=.45\textwidth]{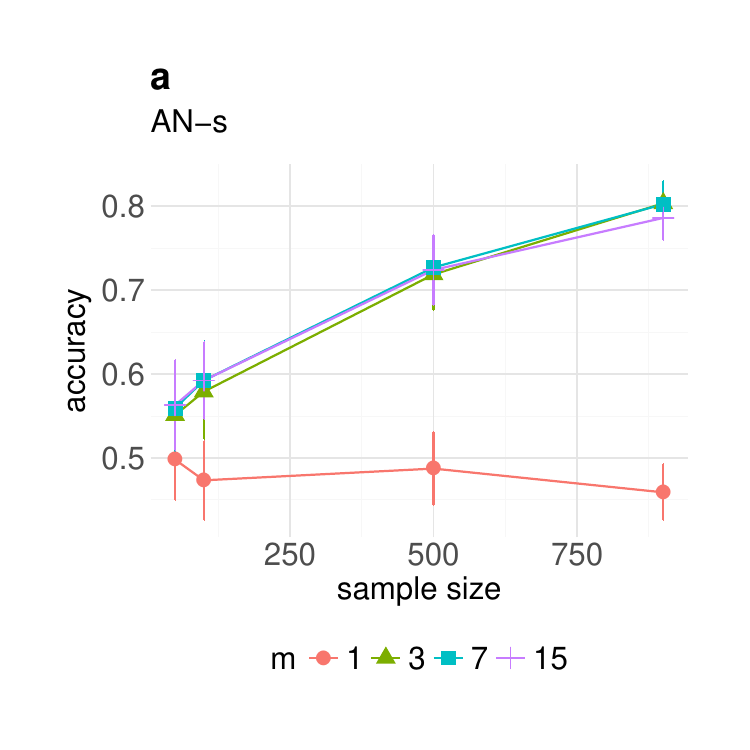} 
 \includegraphics[width=.45\textwidth]{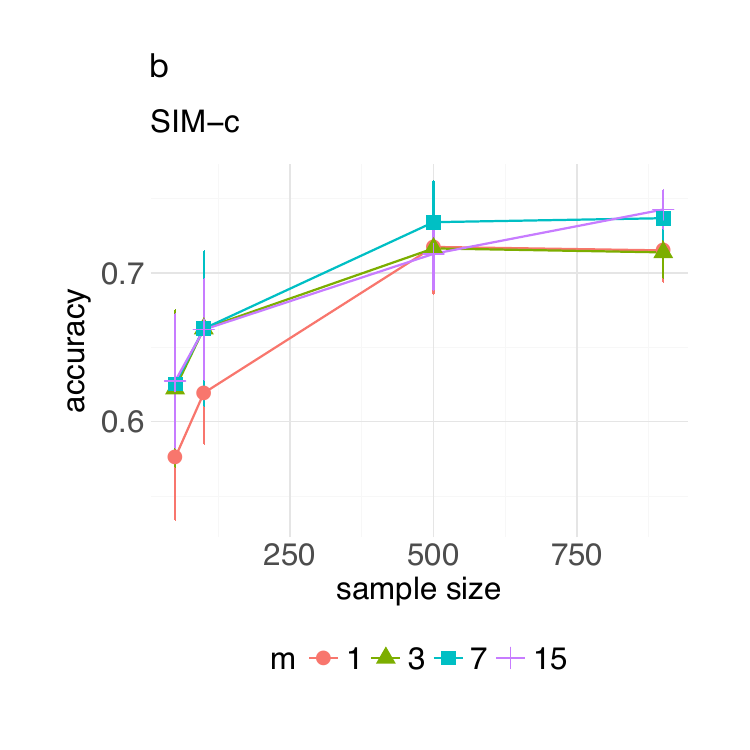}
 }
  \vspace{-1cm}
\caption{The accuracy increases with $m$ and the sample size.}
\label{fig:m_n_roc_dr}
\end{figure}

\textbf{Selection of $m$} 
In an ablation study we explored the significance of the parameter $m$ with regards to different sample sizes. 
Looking at the figures in~\autoref{fig:m_n_roc_dr}, we can clearly see in which cases multiple quantile levels can indeed increase the accuracy, namely for sigmoid (i.e., harder to detect) causal mechanisms.
For confounded data,  increasing $m$ seems to help too, albeit faintlier.
Additionally, we can notice that higher values of $m$ have more pronounced effect as the sample size increases.  
This was used to improve our results on real data, namely by setting the parameter $m$ to 1 when $n < 200$ and $m = 3$ for the rest.

\begin{figure}[h!]
\resizebox{0.45\textwidth}{!}{
\centering
\includegraphics[width=0.7\textwidth]{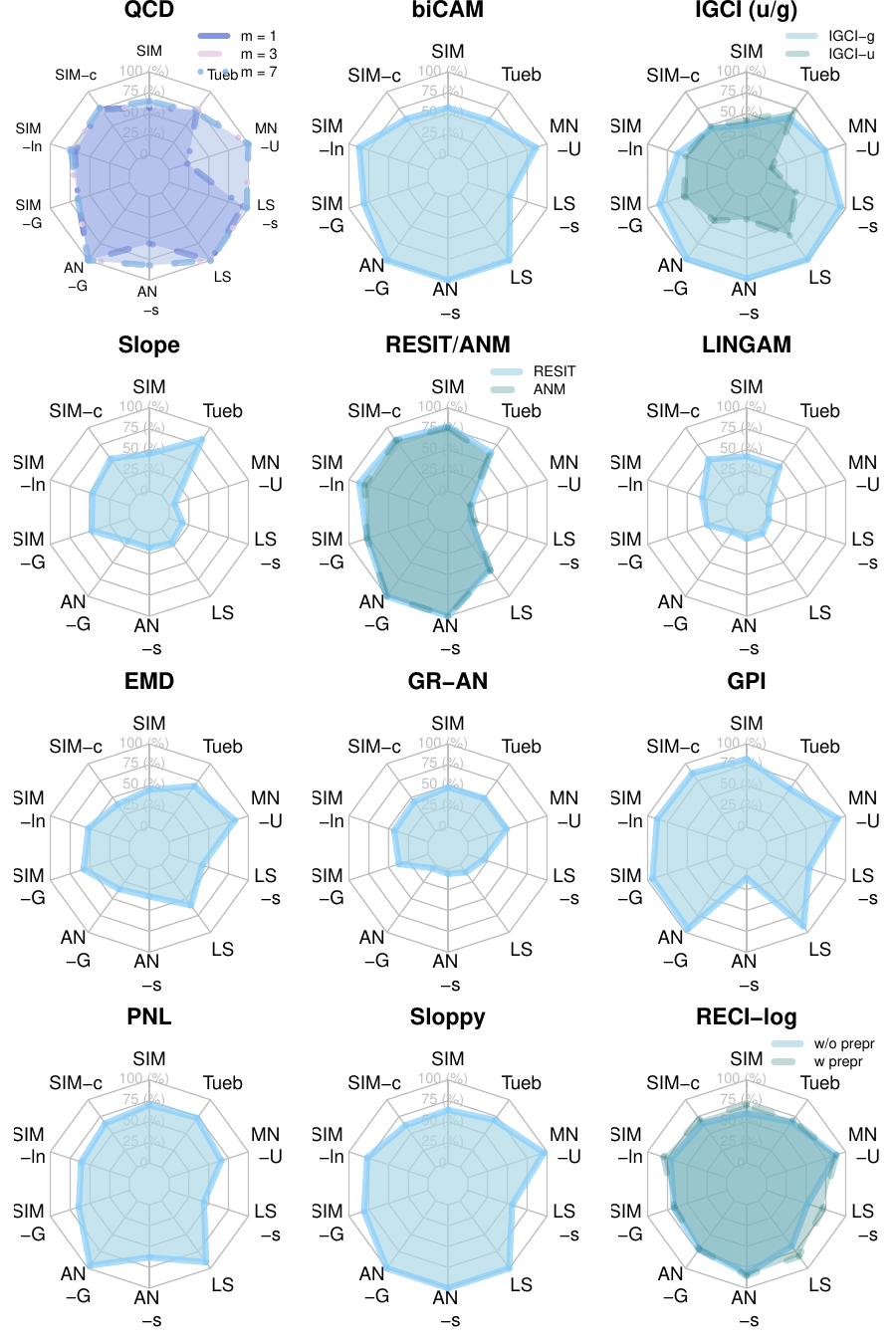}}
\vspace{-0.5cm}
\caption{Accuracy of bQCD and competitors.}
\vspace{-0.45cm}
\label{fig:spider_plots}
\end{figure}

\textbf{Comparison to baselines}
In~\autoref{fig:spider_plots}, we compare causal discovery algorithms across simulated datasets with regards to accuracy.
Tabulated numbers are in \autoref{sec:spider_tab}.

There is no single baseline is an overall best performer, however, we notice that bQCD has most consistent results across all benchmarks.
Starting with the SIM benchmarks, we notice that GPI achieves highest accuracy in all four scenarios, followed with similar results by RESIT/ANM \footnote{Because RESIT is an R version based on the MATLAB ANM, we overlay their results on a single chart.}. 
On this benchmark, bQCD behaves similarly to the rest of the baselines, while being more robust in the confounded scenario where others achieve results on the scale of random guess. 
Interestingly, higher values of the parameter $m$ improve the results in such pairs.

The results are significantly different for the AN, LS and MN scenarios. 
biCAM, ANM and RESIT easily handle the AN pairs since their underlying assumptions are met, while we can notice some discrepancy in the LS and MN scenarios where there is an interaction between the noise and the cause. 
Similarly, LINGAM does not perform well on any of the datasets, which are all highly nonlinear, hence violating its assumptions. 
IGCI can handle any scenario with the gaussian reference measure, while this is not the case with the uniform measure\footnote{Selecting the reference measure is the method's most sensitive part and is as difficult as selecting the right kernel/bandwidth for a specific task \cite{Janzing2012, Mooij2016}.}.
In the LS generative models where not only the mean, but the variance of the effect changes with the cause only IGCI-g was on-par with bQCD, but bQCD is still better than IGCI-g on the SIM benchmark and real data pairs.
On the other hand, more flexible methods such as PNL and EMD had difficulties in the non-injective cases. 
bQCD has satisfactory ($> 75\%$ accuracy) for all different data generative mechanisms (AN, LS and MN). 
From the baselines presented, RECI's model allows for dependence between the noise and the cause.
Although it shows good results in practice, it is important to note that the outcome depends significantly on the preprocessing step as well as the selection of the regressor, for which still, there is no clear guidance. 
For more details on the variations of the proposed solution, we refer to the original paper \citep{blobaum2018cause}.
On the contrary, bQCD achieves the same results no matter the implementation, which makes it straightforward to use.

\begin{figure}
 \resizebox{0.5\textwidth}{!}{
 \includegraphics[width=\textwidth]{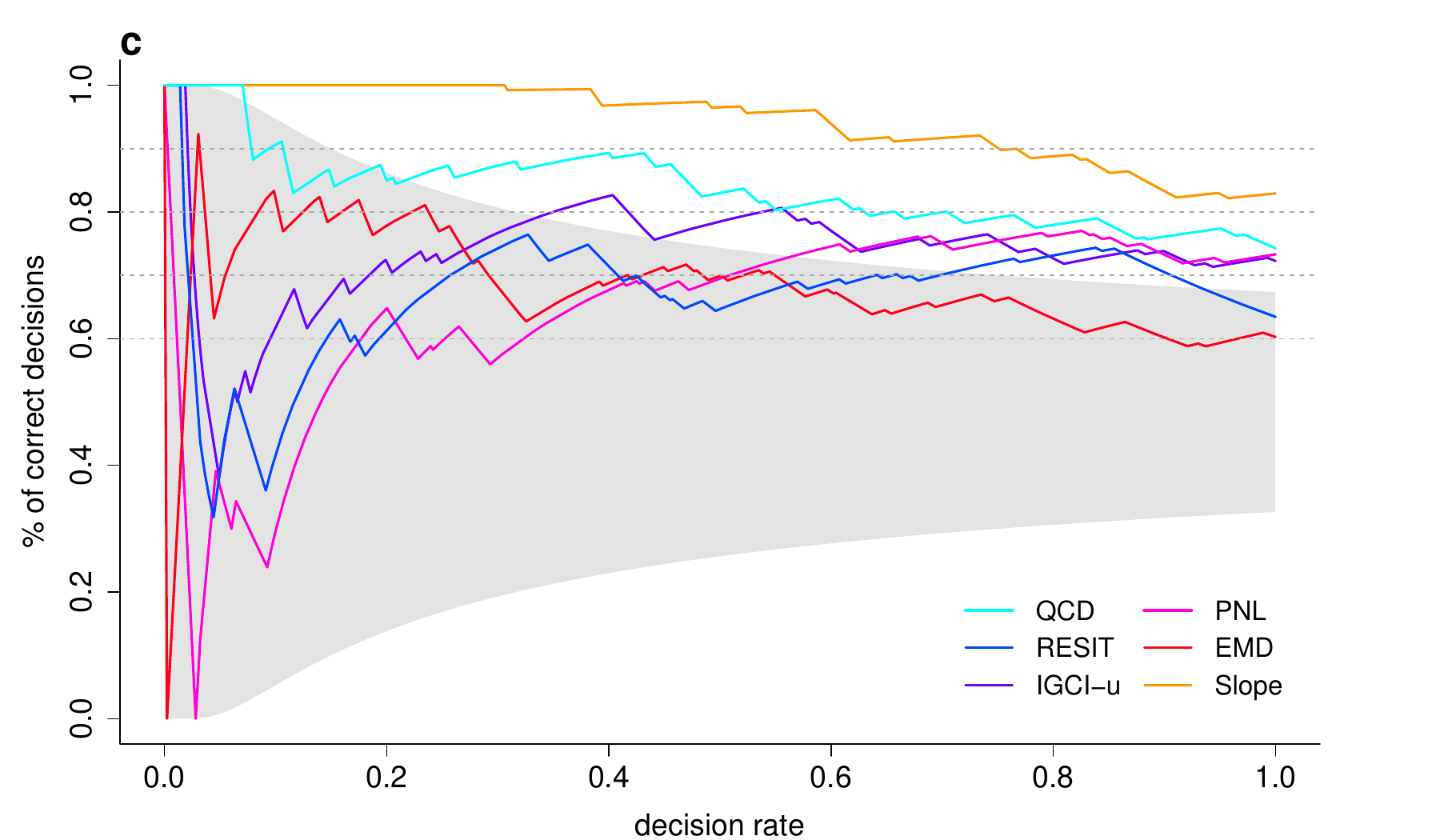}
 }
 \vspace{-0.7cm}
    \caption{Accuracy decision-rates for the top baselines on the T\"{u}bingen Benchmark. }\label{fig:dec_rates}
    \vspace{-0.5cm}
 \end{figure}

With real data pairs,~\autoref{tab:roc_pr_tu} shows that bQCD\footnote{Results are averaged over 30 repetitions to account for the effect of the jittering in the discrete pairs.} is highly competitive in terms of weighted accuracy, with only Slope achieving better overall results. 
However, note that Slope does well only on this dataset and performs poorly on synthetic benchmarks, while bQCD performs well under diverse setups.
Additionally, we include accuracy decision-rate plot in~\autoref{fig:dec_rates}.
Note that bQCD provides statistically significant results (i.e., compared to a coin flip) and is second out of the 6 best performing algorithms with respect to weighted accuracy. 
In \autoref{sec:add_res}, \autoref{fig:apx_roc}, we further provide accuracy decision rate plot and ROC curves with all baselines.
Moreover, the efficiency of our method is highlighted in the last row of~\autoref{tab:roc_pr_tu}, where bQCD is able to go over the whole dataset in $\sim$ 7 minutes.
As for other nonparametric methods, only IGCI and Sloppy are faster, Slope is twice as slow, RESIT 55 times,  PNL 71 times, and the others  required days to go through the whole dataset of had to be averaged on subsamples due to slow execution (GRAN).

Overall, we can conclude that compared to baselines bQCD  performs well in both real and simulated scenarios therefore being more robust to different generative models while also having computational advantages.

\begin{table}
\vspace{-0.5cm}
\label{tab:roc_pr_tu}
 \caption{Results for the T\"{u}bingen Benchmark. }

\resizebox{0.48\textwidth}{!}{

\begin{tabular}{lccccccl}
\hline
 & \textbf{bQCD}     &\textbf{IGCI-u/g}   & \textbf{biCAM}   & \textbf{Slope}   & \textbf{LINGAM}   & \textbf{RESIT} & \textbf{Sloppy}  \\ \hline
\textbf{Acc}       & 0.68           & 0.67/0.61  & 0.57    & 0.75    & 0.3      & 0.53  & 0.59    \\
\textbf{Weig Acc } & 0.75          & 0.72/0.62  & 0.58    & 0.83    & 0.36     & 0.63  & 0.7     \\
\textbf{AUC ROC}   & 0.71          & 0.67       & 0.61    & 0.84    & 0.3      & 0.56  & 0.67    \\
\textbf{CPU}       & 7 min.              & 2 sec.     & 10 sec. & 25 min. & 3.5 sec. & 12 h  & 1.3 min \\ \hline
          & \textbf{EMD} & \textbf{GRAN}       & \textbf{GPI}     & \textbf{PNL-MLP} & \textbf{ANM}      & \textbf{CURE}  & \textbf{RECI}    \\ \hline
\textbf{Acc}       & 0.55                & 0.4  & 0.6     & 0.75    & 0.6      & 0.6   & 0.63    \\
\textbf{Weig Acc}  & 0.6                 & 0.5 & 0.63    & 0.73    & 0.6      & 0.54  & 0.7     \\
\textbf{AUC ROC}   & 0.53                & 0.47   & 0.61    & 0.7     & 0.45     & 0.61  &  0.68       \\
\textbf{CPU}       & 4.6 days            & NA         & 30 days & 8.3 h   & 3.2 days & NA    & 1.2h    \\ \hline
\end{tabular}}
\vspace{-0.5cm}
\end{table}

\section{Conclusion}
\label{sec:conclusion}

In this work, we develop a causal discovery method based on conditional quantiles.
We propose bQCD, an effective implementation of our method based on nonparametric quantile regression, compares favorably to state-of-the-art methods on both simulated and real datasets.
The new method should be preferred mainly because of three reasons.
First, quantiles are less sensitive to outliers than the mean, thus bQCD is robust to contamination or heavy-tails, hence, it does not require preprocesing (``cleaning") step.
Second, we make no assumption on the parametric class considered, thus allowing for a wide range of mechanisms, whereas baselines perform worse when their assumptions are not satisfied.
Third, our implementation is computationally more efficient than other nonparametric methods.
There are currently three directions that we are explorig to extend this work.
First, bQCD can be extended to multivariate cause-effect pairs.
Second, it is interesting to see how quantile scores can fit in the invariant causal prediction framework \citep{Peters2016}.
Third, the computational efficiency of bQCD is promising in the context of extensions to higher dimensional datasets.
As such, ongoing research leverages existing graph discovery algorithms for hybrid learning, as suggested in the supplementary material. 
Fourth, as discussed in \autoref{remark:structural}, finding structural or distributional assumptions for which \autoref{assumption1} or \eqref{eq:cl_to_qs} holds remains an open problem.

\bibliography{icml}
\bibliographystyle{icml2020}

\newpage
\onecolumn
\appendix
\section{Theory}

\subsection{Proof of Theorem 2}
\label{sec:proof_thm2}
We start by proving the sufficiency part of the theorem, namely that, if $\mbox{CL}(l)=o(n)$ for $l \in \{\tau,  q_{X, \tau}, q_{Y, \tau}, q_{Y \vert X, \tau}, q_{X \vert Y, \tau}) \}$, then
\begin{align} \label{eq:pop_cl_conv}
P \left( \frac{\mbox{CL}_{\tau}(X) +  \mbox{CL}_{\tau}(Y \mid X) }  {\mbox{CL}_{\tau}(Y) +  \mbox{CL}_{\tau}(X \mid Y)} \leq 1 \right) \underset{n \to \infty}{\longrightarrow} 1
\end{align}
implies that
\begin{align} \label{eq:qs_ratio_pop}
\frac{\mathbb{E} \left[ S_\tau(q_{X,\tau}, X_i) \right] + \mathbb{E} \left[ S_\tau(q_{Y \mid X = X_i,\tau}, Y_i) \right]}{\mathbb{E} \left[ S_\tau(q_{Y,\tau}, Y_i) \right]+ \mathbb{E} \left[ S_\tau(q_{X \mid Y = Y_i,\tau}, X_i) \right]}\leq 1 .
\end{align}
First, notice that we have
\begin{align} \label{eq:cl_ratio}
\frac{\mbox{CL}_{\tau}(X) +  \mbox{CL}_{\tau}(Y \mid X) }  {\mbox{CL}_{\tau}(Y) +  \mbox{CL}_{\tau}(X \mid Y)}
= \frac{\sum\limits_{i=1}^n S_\tau(q_{X,\tau}, X_i) + \sum\limits_{i=1}^n S_\tau(q_{Y \mid X = X_i,\tau}, Y_i) + K_1 }  {\sum\limits_{i=1}^n S_\tau(q_{Y,\tau}, Y_i) + \sum\limits_{i=1}^n S_\tau(q_{X \mid Y = Y_i,\tau}, X_i) + K_2},
\end{align}
with
\begin{align*}
K_1 &= 2 \mbox{CL}(\tau) + \mbox{CL}(q_{X,\tau}) + \mbox{CL}(q_{Y \mid X,\tau}) - 2 a_n(\tau) = o(n) - 2 a_n(\tau), \\
K_2 &= 2 \mbox{CL}(\tau) + \mbox{CL}(q_{Y,\tau}) + \mbox{CL}(q_{X \mid Y,\tau}) - 2 a_n(\tau) = o(n) - 2 a_n(\tau).
\end{align*}
Thus, we have that
\begin{align} \label{eq:cl_to_qs}
 \frac{\mbox{CL}_{\tau}(X) + \mbox{CL}_{\tau}(Y \mid X)}{\mbox{CL}_{\tau}(Y) + \mbox{CL}_{\tau}(X \mid Y)}  \leq  1 \iff \frac{\sum\limits_{i=1}^n S_\tau(q_{X,\tau}, X_i) + \sum\limits_{i=1}^n S_\tau(q_{Y \mid X = X_i,\tau}, Y_i) }  {\sum\limits_{i=1}^n S_\tau(q_{Y,\tau}, Y_i) + \sum\limits_{i=1}^n S_\tau(q_{X \mid Y = Y_i,\tau}, X_i)} \leq 1 + o_p(1),
\end{align}
where we used the fact that $K_2 - K_1 = o(n)$, which implies that
\begin{align*}
\frac{K_2 - K_1}{\sum\limits_{i=1}^n S_\tau(q_{Y,\tau}, Y_i) + \sum\limits_{i=1}^n S_\tau(q_{X \mid Y = Y_i,\tau}, X_i)} = o_p(1).
\end{align*}
Finally, \eqref{eq:qs_ratio_pop} follows because the law of large numbers along with the continuous mapping theorem imply
\begin{align*} 
\left| \frac{\sum\limits_{i=1}^n S_\tau(q_{X,\tau}, X_i) + \sum\limits_{i=1}^n S_\tau(q_{Y \mid X = X_i,\tau}, Y_i) }  {\sum\limits_{i=1}^n S_\tau(q_{Y,\tau}, Y_i) + \sum\limits_{i=1}^n S_\tau(q_{X \mid Y = Y_i,\tau}, X_i)} - \frac{\mathbb{E} \left[ S_\tau(q_{X,\tau}, X_i) \right] + \mathbb{E} \left[ S_\tau(q_{Y \mid X = X_i,\tau}, Y_i) \right]}{\mathbb{E} \left[ S_\tau(q_{Y,\tau}, Y_i) \right]+ \mathbb{E} \left[ S_\tau(q_{X \mid Y = Y_i,\tau}, X_i) \right]} \right| = o_p(1).
\end{align*}

To prove necessity, assume that
\begin{align*} 
P \left( \frac{\mbox{CL}_{\tau}(Y) +  \mbox{CL}_{\tau}(X \mid Y) }  {\mbox{CL}_{\tau}(X) +  \mbox{CL}_{\tau}(Y \mid X)} \leq 1 \right) \underset{n \to \infty}{\longrightarrow} 1.
\end{align*}
By the sufficiency part of the proof, this implies that
\begin{align*}
\frac{\mathbb{E} \left[ S_\tau(q_{X,\tau}, X_i) \right] + \mathbb{E} \left[ S_\tau(q_{Y \mid X = X_i,\tau}, Y_i) \right]}{\mathbb{E} \left[ S_\tau(q_{Y,\tau}, Y_i) \right]+ \mathbb{E} \left[ S_\tau(q_{X \mid Y = Y_i,\tau}, X_i) \right]} \geq 1.
\end{align*}
which establishes necessity by contraposition.  \hfill $\square$

\subsection{Proof of Theorem 3}
\label{sec:proof_thm3}

We start by proving the sufficiency part of the theorem, namely that, if~\autoref{assumption2} holds, then
\begin{align} \label{eq:pop_cl_conv2}
P \left( \frac{\mbox{CL}_{\tau}(X) +  \mbox{CL}_{\tau}(Y \mid X) }  {\mbox{CL}_{\tau}(Y) +  \mbox{CL}_{\tau}(X \mid Y)} \leq 1 \right) \underset{n \to \infty}{\longrightarrow} 1
\end{align}
implies that
\begin{align} \label{eq:qs_cl_conv}
P\left( \frac{S_{X, \tau}  +  S_{Y \mid X, \tau}}{ S_{Y, \tau}  +  S_{X \mid Y, \tau} }  \leq 1 \right) \underset{n \to \infty}{\longrightarrow} 1.
\end{align}
Denote by $\widehat{\mbox{CL}}_{\tau}(\cdot)$ the quantity corresponding to $\mbox{CL}_{\tau}(X)$ but replacing the population quantile by estimates, namely
\begin{align*}
\widehat{\mbox{CL}}_{\tau}(X)&= \mbox{CL}(\tau) + \mbox{CL}(\widehat{q}_{X,\tau}) + S_{X, \tau} - a_n(\tau), \\
 \widehat{\mbox{CL}}_{\tau}(Y \mid X)&= \mbox{CL}(\tau) + \mbox{CL}(\widehat{q}_{Y \mid X,\tau}) + S_{X \mid Y, \tau}  - a_n(\tau),
\end{align*}
and similarly for $Y$ and $X \mid Y$.
Hence, we have
\begin{align} \label{eq:cl_ratio_hat}
\frac{\widehat{\mbox{CL}}_{\tau}(X) + \widehat{\mbox{CL}}_{\tau}(Y \mid X)}{\widehat{\mbox{CL}}_{\tau}(Y) + \widehat{\mbox{CL}}_{\tau}(X \mid Y)}
= \frac{  S_{X,\tau} + S_{Y \mid X,\tau} + K_1}{S_{Y,\tau} + S_{X \mid Y,\tau} + K_2},
\end{align}
with
\begin{align*}
K_1 &= 2 \mbox{CL}(\tau) + \mbox{CL}(\widehat{q}_{X,\tau}) + \mbox{CL}(\widehat{q}_{Y \mid X,\tau}) - 2 a_n(\tau), \\
K_2 &= 2 \mbox{CL}(\tau) + \mbox{CL}(\widehat{q}_{Y,\tau}) + \mbox{CL}(\widehat{q}_{X \mid Y,\tau}) - 2 a_n(\tau).
\end{align*}
Consistency of the estimators (i.e.,~\autoref{assumption2}) along with the continuous mapping theorem implies
\begin{align} \label{eq:continuous_mapping}
\left| \frac{\mbox{CL}_{\tau}(X) +  \mbox{CL}_{\tau}(Y \mid X) }  {\mbox{CL}_{\tau}(Y) +  \mbox{CL}_{\tau}(X \mid Y)} - \frac{\widehat{\mbox{CL}}_{\tau}(X) + \widehat{\mbox{CL}}_{\tau}(Y \mid X)}{\widehat{\mbox{CL}}_{\tau}(Y) + \widehat{\mbox{CL}}_{\tau}(X \mid Y)} \right| = o_p(1)
\end{align}
Therefore, combining~\eqref{eq:pop_cl_conv2} and~\eqref{eq:continuous_mapping}, we have that
\begin{align} \label{eq:cl_ratio_hat2}
P \left(  \frac{\widehat{\mbox{CL}}_{\tau}(X) + \widehat{\mbox{CL}}_{\tau}(Y \mid X)}{\widehat{\mbox{CL}}_{\tau}(Y) + \widehat{\mbox{CL}}_{\tau}(X \mid Y)} \leq 1 \right) \underset{n \to \infty}{\longrightarrow} 1.
\end{align}
In turn, from~\eqref{eq:cl_ratio_hat}, we have that 
\begin{align} \label{eq:cl_hat_to_qs}
 \frac{\widehat{\mbox{CL}}_{\tau}(X) + \widehat{\mbox{CL}}_{\tau}(Y \mid X)}{\widehat{\mbox{CL}}_{\tau}(Y) + \widehat{\mbox{CL}}_{\tau}(X \mid Y)}  \leq  1 \iff \frac{  S_{X,\tau} + S_{Y \mid X,\tau}}{S_{Y,\tau} + S_{X \mid Y,\tau}} \leq 1 + \frac{K_2 - K_1}{S_{Y,\tau} + S_{X \mid Y,\tau}} = 1 + o_p(1),
\end{align}
where the last equality follows from assuming, without loss of generality, that $CL(\tau) = o(n)$, which implies that $K_1 = o(n) - 2 a_n(\tau)$ and $K_2 = o(n) - 2 a_n(\tau)$ and thus that $K_2 - K_1 = o(n)$.
Finally,~\eqref{eq:qs_cl_conv} follows from combining~\eqref{eq:cl_ratio_hat2} and~\eqref{eq:cl_hat_to_qs}.

To prove necessity, assume that
\begin{align*} 
P \left( \frac{\mbox{CL}_{\tau}(Y) +  \mbox{CL}_{\tau}(X \mid Y) }  {\mbox{CL}_{\tau}(X) +  \mbox{CL}_{\tau}(Y \mid X)} \leq 1 \right) \underset{n \to \infty}{\longrightarrow} 1.
\end{align*}
By the sufficiency part of the proof, this implies that
\begin{align*}
P \left( \frac{  S_{X,\tau} + S_{Y \mid X,\tau}}{S_{Y,\tau} + S_{X \mid Y,\tau}} \geq 1 \right) \underset{n \to \infty}{\longrightarrow} 1.
\end{align*}
which establishes necessity by contraposition.  \hfill $\square$

\vspace{0.5cm}

\section{bQCD implementation}
\label{sec:quant_copula_details}

To leverage \autoref{thm:qs_assym} and \autoref{thm:causal_rule} for causal discovery, 
we need a flexible model that yields computable expressions for all conditional quantiles. 
We show the efficiency of our method independent of the choice of underlying regressor through various implementations, namely copulas, quantile forests and neural networks.

\begin{minipage}{0.75\textwidth}%
\begin{algorithm}[H] 
\caption{bQCD algorithm}
\begin{algorithmic}
\STATE {\bfseries Inputs:} \\
\begin{itemize}
\item  $\left\{X_i,Y_i \right\}^n_{i=1}$, $n$ i.i.d. observations of the rvs $X$ and $Y$,
\item $qr$, a quantile regression method,
\item $m$, the number of quantiles levels (weights/nodes pairs to approximate the integral).
\end{itemize}
\STATE 1. Transform  $X$ and $Y$ to standard normal.
\STATE 2. Compute the quadrature weights and nodes $\left\{ w_j, 
\tau_j\right\}_{j=1}^m$.
\STATE 3. Compute the conditional and marginal scores for $X \rightarrow Y$ (and similarly for $Y \rightarrow X$):
\FOR{$j=1$ {\bfseries to} $m$}
\STATE 
\begin{itemize}
\item Use the empirical quantile to compute $\widehat{q}_{X,\tau_j}$ and  $S_{X, \tau_j} = \sum_{i=1}^n S_{\tau_j}(\widehat{q}_{X,\tau_j}, X_i)$.
\item Use the method $qr$ to compute $\widehat{q}_{X | Y, \tau_j}$ and $S_{Y \mid X, \tau_j} = \sum_{i=1}^n S_{\tau_j}(\widehat{q}_{Y \mid X = X_i,\tau_j}, Y_i)$.

\end{itemize}
\ENDFOR
\STATE 4. Compute the scores
 $\widehat{S}_{\star} = \sum_{j=1}^m w_j S_{\star,\tau_j}$ for $\star \in \{X, Y, Y \mid X, X \mid Y\}$.
\STATE 5. Define $S_{X \to Y} = \widehat{S}_{X} + \widehat{S}_{Y \vert X}$ and $S_{Y \to X} = \widehat{S}_{Y} + \widehat{S}_{X \vert Y}$.
\IF{$S_{Y \to X} > S_{X \to Y}  $}
\STATE Define the causal direction as $d = X \rightarrow Y$.
\ELSIF{$S_{Y \to X} < S_{X \to Y}  $}
\STATE Define the causal direction as $d = Y \rightarrow X$.
\ENDIF
\STATE  Define the confidence score as $C = -S_{X \to Y} + S_{Y \to X}$.
\STATE {\bfseries Output:} ($d$, $C$)
\end{algorithmic}
\end{algorithm}\label{algo}
\end{minipage}

\paragraph{Nonparametric copulas} Copulas (i.e., multivariate distributions with uniform margins), represent an appealing alternative to compute conditional quantiles. 
In a nutshell, copulas make it possible to flexibly model the dependence structure between random variables while avoiding assumptions about the scales, functional forms, or other restrictions imposed when dealing with marginal distributions.

Following \cite{joe1996}, copulas lead to a useful representation of the distributions of $Y \vert X$ and $X \vert Y$, namely $F_{Y \vert X}(x,y) = P(Y \leq y \vert X = x) = P(V \leq v | U = u) = \partial_u C(u, v)$ where $C$ is a copula, $\partial_u C(u,v) = \partial C(u,v)/\partial u$, $u= F_X(x)$, and 
$v=F_Y(y)$, and similarly $F_{X \vert Y}(x,y) = \partial_v C(u,v)$.
This means that conditional distributions can be evaluated by taking partial derivatives of the copula function.
The $\tau$-quantiles of $Y \vert X$ can then be written as $
F^{-1}_{Y \vert X}(x,\tau) = F^{-1}_{Y}((\partial_u C)^{-1}(u, \tau))$
and similarly for $F^{-1}_{X \vert Y}(\tau,y)$, with inverse functions of the copula derivatives are with respect to $\tau$.
Using the above equations, one can then compute all $\tau$-quantiles using the marginal distributions and the copula.
To benefit from a fully nonparametric estimation, in practice we use nonparametric copulas \citep{Geenens2017}.
We implemented Quantile Copula Causal Discovery (bQCD) using the R interface \cite{R} interface to \texttt{vinecopulib} called \texttt{rvinecopulib} \cite{rvinecopulib} and Gauss-Legendre quadrature from the package \texttt{statmod} \cite{statmod}.
We refer to \autoref{sec:quant_copula} for more details.

\paragraph{Quantile forest} 
Another nonparametric and consistent way of estimating conditional quantiles is via quantile regression forests.
In contrast to random forests \citep{breiman2001random}, for each node in each tree, quantile regression forest keeps the value of all observations in this node, not just their mean, which allows us to assess the conditional distribution based on this information.
For detailed explanation of the algorithm and its consistency we refer the reader to the original paper \cite{meinshausen2006quantile}.
In experiments we used \verb|nodesize=10|, \verb| sampsize=50| and \verb|n_trees=500| for both regression directions.

\paragraph{Quantile neural network}
Recently, quantile estimation has become of interest in the deep learning community as part of uncertainty estimation framework.
We leverage such an approach for a flexible approximation of the conditional quantiles which are estimated simultaneously by a neural network trained with pinball loss, i.e. the quantile scoring as a loss function.
In the experiments we use the \verb|qrnn| package by \citet{Cannon2018} which includes monotonicity constraints to guard against quantile crossing.
Again, for both sides (possible causal directions) we use symmetric models, which is a two-layered neural network with 10 neurons per layer, with sigmoid nonlinearities.
The networks were trained to 1000 iterations.

\begin{figure}[H]
\centering
\resizebox{0.7\textwidth}{!}{
 \includegraphics{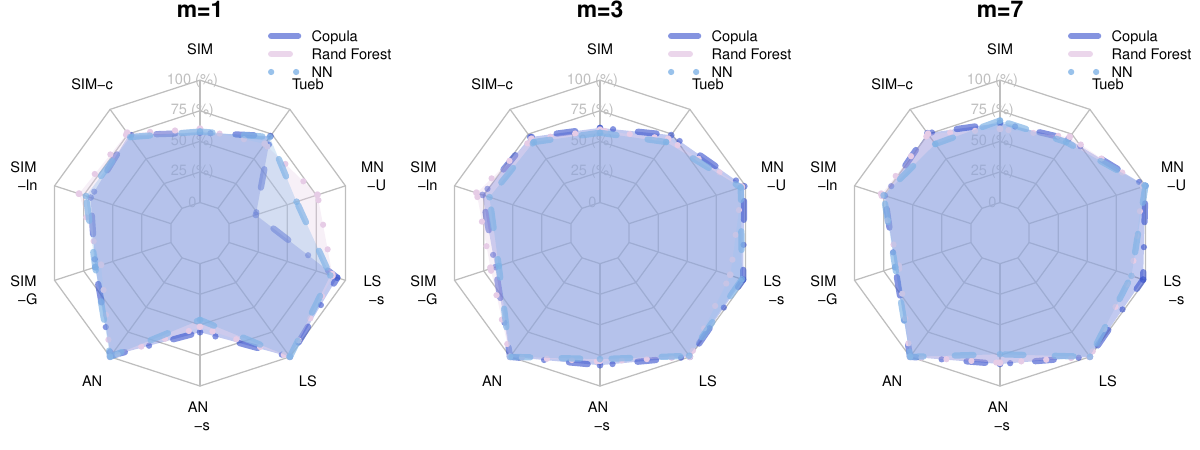}
 }
\caption{bQCD robustness with respect to implementation for different number quantiles.}
\label{fig:bQCD_all_m}
\end{figure}

\subsection{Predicting Quantiles with Copulas}
\label{sec:quant_copula}
While not the only statistical model satisfying \autoref{assumption2}, copulas (i.e., multivariate distributions with uniform margins), represent an appealing alternative.
Any bivariate nonparametric estimator allowing to compute the required inverse conditional CDFs could be used.
But, to the best of our knowledge, no available implementation allows directly for this.
With copulas, obtaining conditional quantiles is straightforward through \eqref{eq:condquant}.
And they are implemented efficiently because they appear in higher-dimensional models such as vines \citep{bedford2002vines, aas2009pair}.

According to the theorem of \cite{sklar1959}, any $F$ can be represented by its marginal distributions $F_X, F_Y$ and a copula $C$, which is is the joint distribution of $(U, V) = \bigl(F_X(X), F_Y(Y)\bigr)$.
In other words, for any $F$, there exists a $C$ such that $F(x,y)=C(F_X(x),F_Y(y))$ for each $(x,y) \in \R^2$. 
Moreover, if all the distributions are continuous, then $C$ is unique.

Following \cite{joe1996}, copulas lead to a useful representation of the distributions of $Y \vert X$ and $X \vert Y$, namely $F_{Y \vert X}(x,y) = P(Y \leq y \vert X = x) = P(V \leq v | U = u) = \partial_u C(u, v)$ where $\partial_u C(u,v) = \partial C(u,v)/\partial u$, $u= F_X(x)$, and 
$v=F_Y(y)$, and similarly $F_{X \vert Y}(x,y) = \partial_v C(u,v)$.
This means that conditional distributions can be evaluated by taking partial derivatives of the copula function.
The $\tau$-quantiles of $Y \vert X$ can then be written as
\begin{align}\label{eq:condquant}
F^{-1}_{Y \vert X}(x,\tau) = F^{-1}_{Y}((\partial_u C)^{-1}(u, \tau)),
\end{align}
and similarly for $F^{-1}_{X \vert Y}(\tau,y)$, with inverse functions of the copula derivatives are with respect to $\tau$.
Using~\eqref{eq:condquant}, one can then compute all $\tau$-quantiles using the marginal distributions and the copula.

Assume that we have $n$ i.i.d. random variables $\left\{(X_1, Y_1), \cdots, (X_n, Y_n) \right\}$.
To avoid relying on restrictive assumptions, $S_{X \to Y}$ and $S_{Y \to X}$ are estimated completely nonparametrically.

Note that, if all the considered distributions are differentiable, $F(x,y)=C(F_X(x),F_Y(y))$ implies
\begin{align}\label{eq:density}
f(x,y) = c\bigl\{ F_X(x), F_Y(y)\bigr\} f_X(x) f_Y(x),
\end{align}
where $f,c, f_X$, and $f_Y$ are the densities corresponding to $F,C, F_X$, and 
$F_Y$ respectively. 
Equation~\eqref{eq:density} has an important implication for inference: because the right-hand side is a product, the joint log-likelihood can be written as a  sum of the log-likelihood of each margin and the log-likelihood of the copula.
This fact can be conveniently exploited in a two-step procedure: estimate the margins separately to obtain $\widehat{F}_X$ and 
$\widehat{F}_Y$, and then take the probability integral transform of the data using those margins, 
that is, define $\widehat{U}_i=\widehat{F}_X(X_i)$ and 
$\widehat{V}_i=\widehat{F}_Y(Y_i)$, to estimate $\widehat{C}$.

For the first step, we simply use the empirical distribution $\widehat{F}_X(x)=\sum_{i=1}^n \mathbb{I} \left\{X_i \leq x \right\}/(n+1)$ (and similarly for $Y$), where $n+1$ is used instead of $n$ in the copula context to avoid boundary problems.
Discrete datasets are handled by \emph{jittering}, that is breaking ties at random.
As for the second step, since typical nonparametric estimators are targeted at densities with unbounded support, they are unsuited to densities restricted to $[0,1]^2$.
To get around this issue, \cite{scaillet2007} suggests to first transform the data to standard normal margins, and then use any nonparametric estimator suited to unbounded densities.
The transformation estimator of the copula density $c(u,v)$ is then defined as
\begin{align}\label{eq:tll}
\widehat{c}(u,v)=
\frac{\widehat{g}(\Phi^{-1}(u) - \Phi^{-1}(\widehat{U}_i),\Phi^{-1}(v) - 
\Phi^{-1}(\widehat{V}_i))}{\phi (\Phi^{-1}(u)) \phi (\Phi^{-1}(v))},
\end{align}
where $\widehat{U}_i = \widehat{F}_X(X_i)$, $\widehat{V}_i = \widehat{F}_Y(Y_i)$, $\widehat{g}$ is a bivariate nonparametric estimator, and $\phi, \Phi$ denote the standard normal density and distribution, respectively.
Following \cite{scaillet2007,Geenens2017}, we then use a bivariate Gaussian kernel estimator for $\widehat{g}$.

The consistency and asymptotic normality of this estimator are derived in \cite{Geenens2017}.
Given that the empirical distribution is also consistent, consistency of the resulting joint distribution is also straightforward.
Furthermore, using the fact that $F_{X}(x) = F(x, \infty)$, $F_{Y}(y) = F(\infty, y)$, $F_{X \vert Y}(x, y) =\partial_y F(x, y)/\left. \partial_y F(x, y) \right|_{x \to \infty}$, $F_{Y \vert X}(x, y) = \partial_x F(x, y)/\left. \partial_x F(x, y) \right|_{y \to \infty}$, consistent estimators of the marginal and conditional distributions are obtained by replacing $F$ by the copula-based joint estimator and using the same relations.

The transformation kernel estimator for bivariate copula densities is implemented in \texttt{C++} as part of \texttt{vinecopulib} \cite{vinecopulib}, a header-only \texttt{C++} library for copula models based on \texttt{Eigen} 
\cite*{eigenweb} and \texttt{Boost} \cite{Schaling2011}.
From~\eqref{eq:tll}, \texttt{vinecopulib} constructs and stores a $30 \times 30$ grid over $[0,1]^2$ along with the evaluated density at the grid points\footnote{
After extensive simulations comparing copula densities to their kernel approximations, the authors of \texttt{vinecopulib} noticed that the precision gains achieved by increasing beyond  $30 \times 30$  were marginal. 
Using a grid is not actually required to implement the kernel-based estimator, but evaluating kernels for each call to the copula's density is computationally expensive and storing the values is a time-memory trade-off.}.
Then, a cubic-spline approximation makes it possible to efficiently compute the copula distribution $\widehat{C}(u,v)$ and its derivatives,   $\partial_u\widehat{C}(u,v)$ and $\partial_v \widehat{C}(u,v)$, as the integrals of the spline-approximation of the density admits an analytic expression.
Finally, \texttt{vinecopulib} implements the numerical inversion of $\partial_u\widehat{C}(u,v)$ and $\partial_v \widehat{C}(u,v)$ to compute  $(\partial_u\widehat{C})^{-1}(u,v)$ and $(\partial_v\widehat{C})^{-1}(u,v)$ by a vectorized version of the bisection method.

\vspace{0.5cm}

\section{Baseline implementation details}

\subsection{Baselines - Implementations and hyper parameters} 
\label{sec:impl_details}
For IGCI, we use the original implementation of \citet{Janzing2012} with 
slope-based estimation with both gaussian and uniform reference measure. 
For LINGAM, we use the implementation of \citet{Peters2014}, which also provides 
RESIT with GP regression and the HSIC independence test with a threshold value 
$\alpha = 0.05$. 
For CAM, we use the R package \citep{Peters2015b} with the default parameters. 
For GR-AN and EMD, we use the code of \citet{Hernandez-Lobato2016a}. 
The parameters for EMD on simulated data are as in the original paper, $\lambda 
= 1e^{-3}$ and $\sigma = \frac{1}{5} S_m$ ($S_m$ being the median of distances 
across all input patterns).
For the real data benchmark, GR-AN was evaluated over 15 subsamples limited to 500 observations, and the results are then averaged.
However, the parameters are tuned and selected from the overall best results. 
GR-AN has a built-in function that takes care parameter tuning. 

For Slope, we use the implementation of \citet{Marx2017}, with local regression 
included in the fitting process. 
For PNL, GPI-MML, and ANM, we use the MATLAB implementation from the 
\href{https://github.com/ssamot/causality}{ \emph{Cause Effect Pairs Challenge 
FirfiD}}, with the same setup as RESIT for ANM. 
For Sloppy we use the results as provided with the code of \citep{marx2019identifiability} with the AIC score.
For RECI we use an example MATLAB code provided by the authors with the 'log' implementation, with and without the data preprocessing step.
Because there is no publicly available implementation of CURE, we use the 
results from files obtained by  \citet{Marx2017}.
Our code and datasets used for the paper are included in the submitted supplementary package.

All experiments are carried out using R version 3.4.1 on a MacBook Pro 
(mid-2015) with 16GB memory, a 2.2 GHz Intel Core i7 processor and 250GB SSD.

\subsection{Computational complexity}
\label{sec:complexity}
\begin{figure}[H]
\centering
\includegraphics[width=0.4\textwidth]{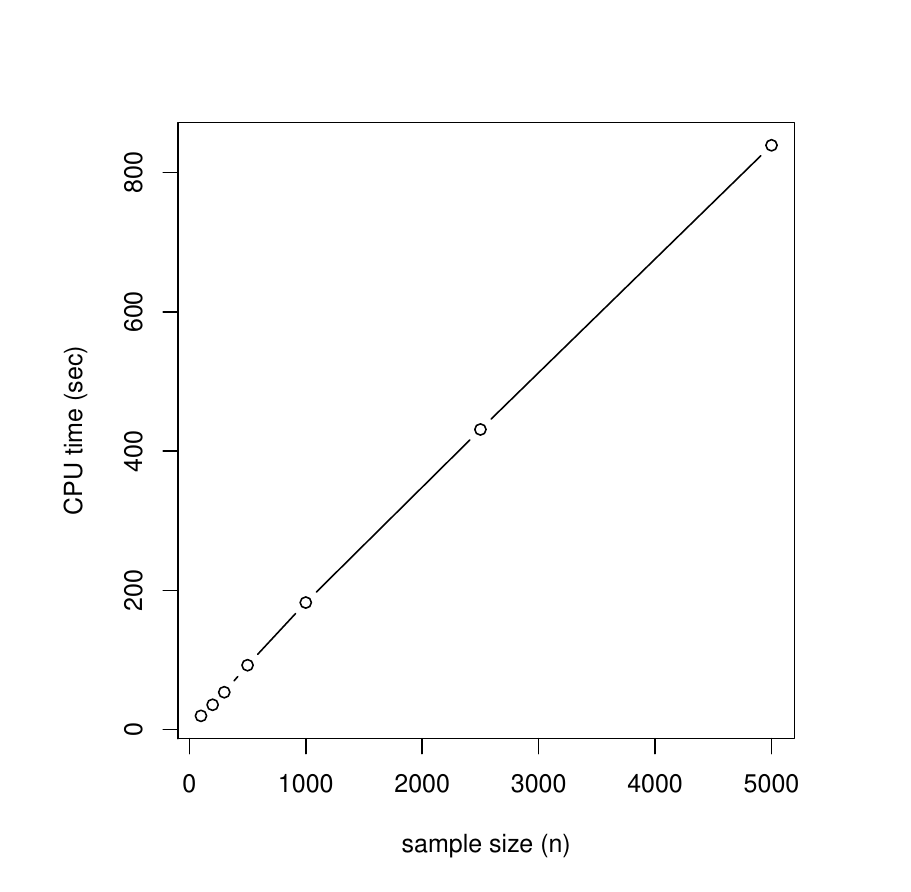}
\caption{Linear scaling of bQCD over 100 pairs, $m = 1$, $n$ varies}
\label{fig:comp_complexity}
\end{figure}

\section{Additional results from experiments}
\label{sec:add_res}

\subsection{Tabular results for spider plots}\label{sec:spider_tab}
\begin{table}[H]
\caption{Accuracy on simulated and real data.}
\label{tab:simulations}
\small
\resizebox{\textwidth}{!}{
\begin{tabular}{@{}lccccccccccc@{}}
\toprule
\textbf{}           & \textbf{SIM} & \textbf{SIM-c} & \textbf{SIM-ln} & \textbf{SIM-G} & \textbf{AN} & \textbf{AN-s} & \textbf{LS} & \textbf{LS-s} & \textbf{MN-U} & \textbf{T\"{u}b} & \textbf{T\"{u}b weigh.} \\ \midrule
\textbf{bQCD m = 1} & 0.57          & 0.69           & 0.69            & 0.59           & 0.99        & 0.47          & 1         & 0.93          & 0.23          & 0.64                               & 0.74                                     \\
\textbf{bQCD m = 3} & 0.61         & 0.72           & 0.76            & 0.65           & 1           & 0.81          & 1           & 0.98          & 0.99          & 0.69                               & 0.77                                     \\
\textbf{bQCD m = 7} & 0.65         & 0.76           & 0.74            & 0.66           & 1           & 0.82          & 1           & 0.98          & 0.99          & 0.67                               & 0.75                                     \\
\textbf{IGCI - u}   & 0.42         & 0.49           & 0.52            & 0.54           & 0.41        & 0.27          & 0.63        & 0.37          & 0.07          & 0.67                               & 0.72                                     \\
\textbf{IGCI - g}   & 0.36         & 0.46           & 0.62            & 0.85           & 0.98        & 0.98          & 0.99        & 0.94          & 0.74          & 0.61                               & 0.64                                     \\
\textbf{biCAM}      & 0.57         & 0.6            & 0.87            & 0.81           & 1           & 1             & 1           & 0.53          & 0.86          & 0.57                               & 0.58                                     \\
\textbf{Slope}      & 0.45         & 0.54           & 0.47            & 0.48           & 0.18        & 0.18          & 0.21        & 0.17          & 0.07          & 0.75                               & 0.83                                     \\
\textbf{RESIT}      & 0.78         & 0.82           & 0.87            & 0.77           & 1           & 1             & 0.6         & 0.03          & 0.05          & 0.53                               & 0.63                                     \\
\textbf{ANM}        & 0.76         & 0.81           & 0.80            & 0.77           & 1           & 1             & 0.62        & 0.09          & 0.03          & 0.59                               & 0.59                                     \\
\textbf{LINGAM}     & 0.42         & 0.53           & 0.31            & 0.25           & 0           & 0.04          & 0.07        & 0.03          & 0             & 0.27                               & 0.42                                     \\
\textbf{EMD}        & 0.45         & 0.4            & 0.52            & 0.58           & 0.36        & 0.33          & 0.6         & 0.42          & 0.83          & 0.6                                & 0.68                                     \\
\textbf{GRAN}       & 0.48         & 0.44           & 0.43            & 0.37           & 0.05        & 0.06         & 0.11        & 0.20          & 0.5           & 0.4                                & 0.5                                      \\
\textbf{GPI}        & 0.82         & 0.86           & 0.88            & 0.94           & 0.95        & 0.11          & 0.91        & 0.54          & 0.90          & 0.6                                & 0.63                                     \\
\textbf{PNL}        & 0.70         & 0.65           & 0.61            & 0.64           & 0.96        & 0.63          & 0.91        & 0.44          & 0.66          & 0.75                               & 0.73                                     \\
\textbf{Sloppy}        & 0.64	 & 0.62	 & 0.77 	& 0.81 & 1	& 1 	& 1 	& 		0.56 	 & 0.96, 	& 0.59  & 0.74                                    \\ 
\textbf{RECI w/wo pp}        & 0.7/0.6	 & 0.71/0.65	 & 0.8/0.71	& 0.66/0.67 & 0.72/0.72	& 0.15/0.17 	& 0.82/0.67 	& 		0.71/0.52 	 & 0.88/0.88	& 0.61/0.63  & 0.7/0.67                                  \\ \bottomrule
\end{tabular}}
\end{table}

\begin{table}[H]
\tiny
\centering
\caption{Accuracy on simulated and real data for different selection of $\tau$ values.}
\resizebox{\textwidth}{!}{
\begin{tabular}{@{}lccccccccccc@{}}
\toprule
\textbf{}                                             & \multicolumn{1}{l}{\textbf{SIM}} & \multicolumn{1}{l}{\textbf{SIM-c}} & \multicolumn{1}{l}{\textbf{SIM-ln}} & \multicolumn{1}{l}{\textbf{SIM-G}} & \multicolumn{1}{l}{\textbf{AN}} & \multicolumn{1}{l}{\textbf{AN-s}} & \multicolumn{1}{l}{\textbf{LS}} & \multicolumn{1}{l}{\textbf{LS-s}} & \multicolumn{1}{l}{\textbf{MN-U}} & \multicolumn{1}{l}{\textbf{Tueb}} & \multicolumn{1}{l}{\textbf{Tueb-w}} \\ \midrule
\textbf{$\tau \in [0.1, 0.5, 0.9]$} & 0.61                             & 0.72                               & 0.76                                & 0.65                               & 1                               & 0.81                              & 1                               & 0.98                              & 0.99                              & 0.69                              & 0.77                                \\
\textbf{$\tau \in [0.4, 0.5, 0.6]$}   & 0.56                             & 0.71                               & 0.72                                & 0.58                               & 1                               & 0.45                              & 1                               & 0.95                              & 0.25                              &   0.63                                & 0.73                               \\ \bottomrule
\end{tabular}}
\end{table}

\begin{table}[H]
\caption{Area under the ROC and PR curves for bQCD.}
\label{tab:roc_pr}
\tiny
\resizebox{\textwidth}{!}{
\begin{tabular}{llclcccccc}
\hline
        & \textbf{SIM} & \textbf{SIM-c} & \textbf{SIM-ln} & \textbf{SIM-G} & \textbf{AN} & \textbf{AN-s} & \textbf{LS} & \textbf{LS-s} & \textbf{MN-U} \\ \hline
ROC-AUC & 0.67         & 0.79           & 0.87            & 0.69           & 1           & 0.9           & 1           & 1             & 1             \\
PR-AUC  & 0.67         & 0.82           & 0.86            & 0.7            & 1           & 0.89          & 1           & 1             & 1             \\ \hline
\end{tabular}}
\end{table}

\begin{figure}[H]     
   \centering
            \includegraphics[width=.3\textwidth]{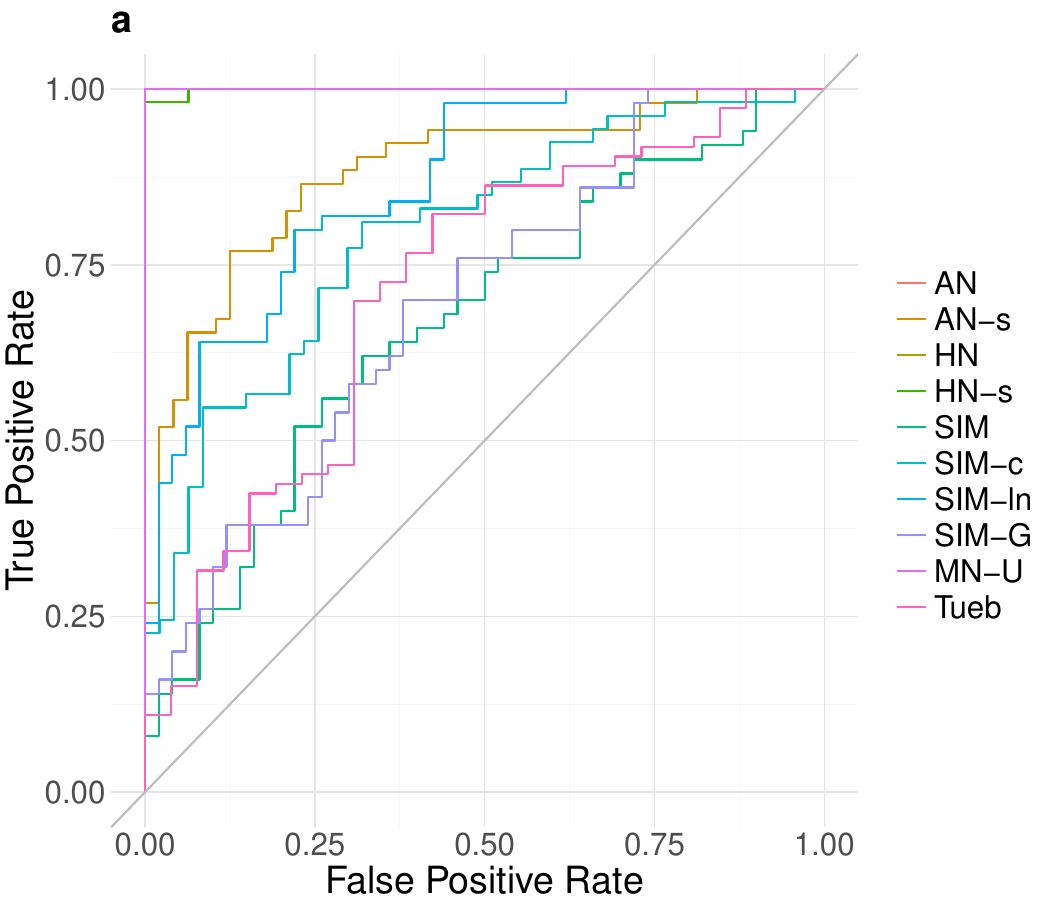}
           \includegraphics[width=0.3\columnwidth]{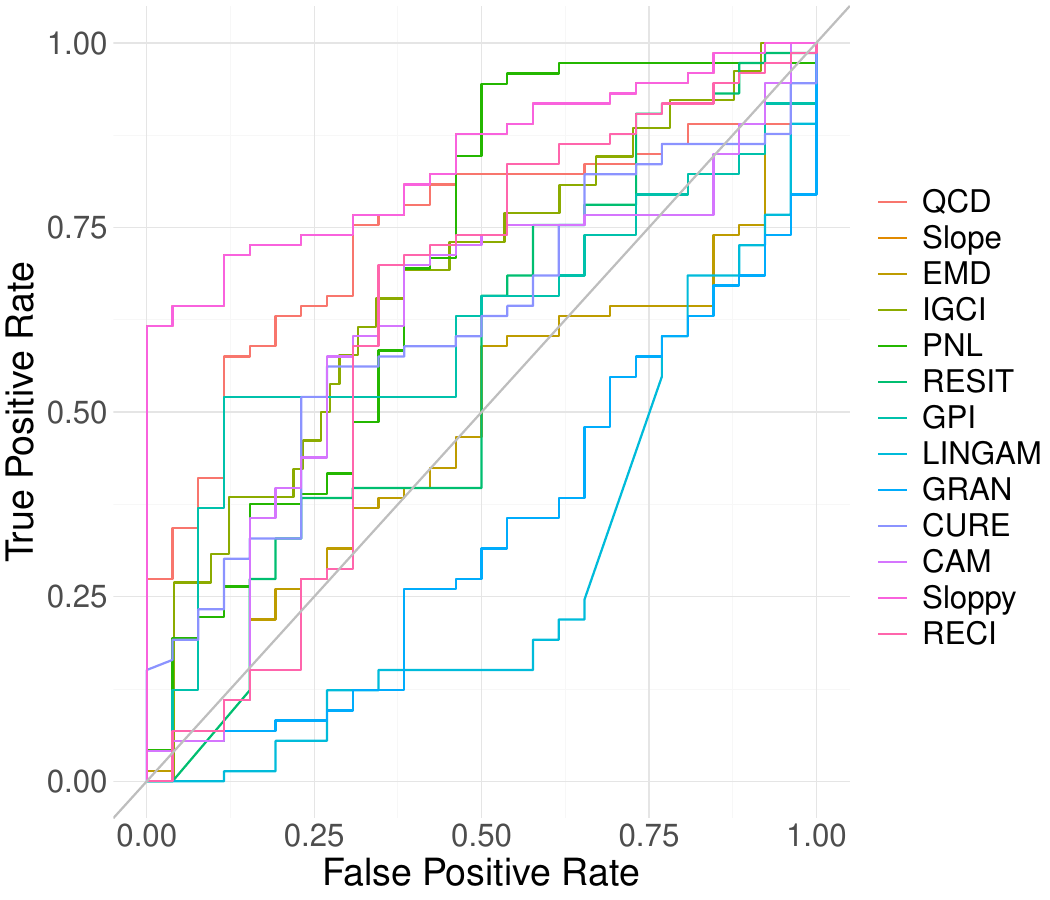}
           \includegraphics[width=0.3\columnwidth]{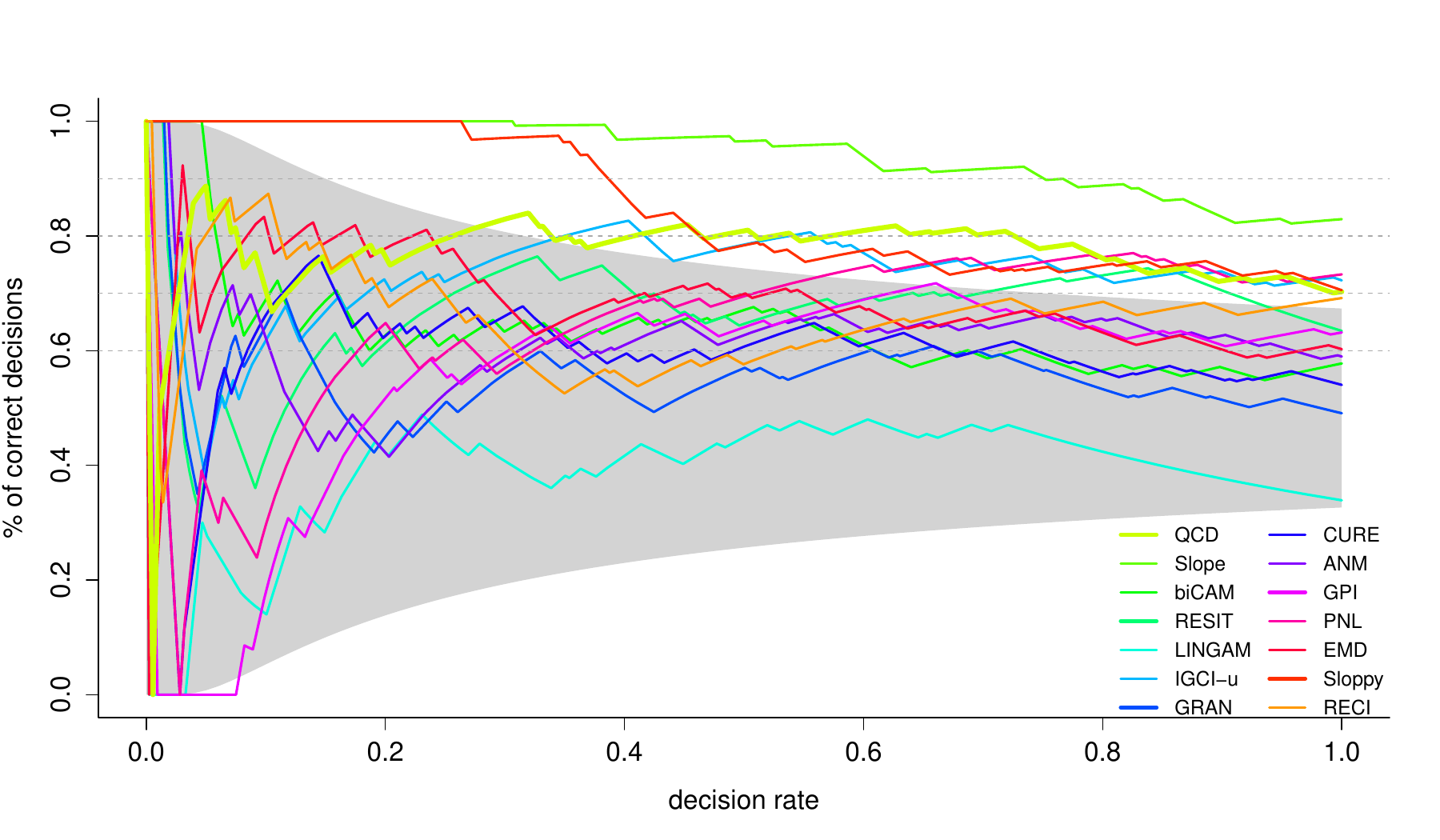}       
\caption{ROC curves  (a) for bQCD all benchmarks, (b) all baselines on Tuebingen, (c) and weighted accuracy-decision rate  curves for all baselines on the 
T\"{u}bingen dataset.}\label{fig:apx_roc}
        \end{figure}

Note that due to RESIT being undecided in few of the real data pairs, we observe the random classifier behavior at the beginning of its ROC curve in \autoref{fig:apx_roc}.

\vspace{-0.5cm}

\vspace{-1.5cm}

\section{Extensions of the pairwise method}
\label{sec:dags}

Recent work by \citet{Goudet2018} suggests that pairwise and CPDAG (i.e., the skeleton 
and the $v$-structures) learning procedures can suitably 
complement each other.
In this section, we follow a similar approach to suggest an extension of bQCD to 
multivariate datasets:
start from the CPDAG resulting from another method,
and then use bQCD to orient the edges.
We rank edges and include them in the graph sequentially, starting from 
the edge with the highest confidence, while checking the acyclicity of the 
resulting graph after each addition.
Note that this approach also requires a final verification to test edge 
orientation and $v$-structures consistency \citep{Goudet2018}. 

We do this by using CAM to learn a CPDAG and then use bQCD to orient its edges.
The rationale is that, while \citet{Buhlmann2014} proves the consistency of the variable ordering by CAM 
used to learn the structure even when its assumptions are not met, bQCD is better in pairwise 
discoveries (see \autoref{sec:experiments}).
But we could also have used the PC algorithm \citep{Spirtes2000}.
 
 We now turn to the well know protein causal dataset by 
\citet{Sachs2005}, 
 for which a ground truth DAG of the causal structure is provided. We used the 
\emph{cd3cd28}
 dataset with 853 observations of 11 proteins. 
For evaluation we use the structural hamming distance (SHD) as 
proposed in \citep{Tsamardinos2006} (adding, removing or reversing an edge) 
necessary to 
transform one graph to the another and the structural intervention distance 
(SID), as in \citep{Peters2015}, which is considered to
be appropriate for quantifying the correct order among variables, by estimating 
the causal effects they entail.
CAM by itself outputs a DAG with $SID_{CAM} = 53$, and  $SHD_{CAM} = 17$, while 
with bQCD we can reduce this to average
 $SID_{bQCD} = 46 (2.5)$, and average $SHD_{bQCD} = 15.1(1.6)$. 
 In \autoref{tab:sachs} we provide the average results over ten random sub-samples of the dataset where the edges of CAM's CPDAG were oriented by each of the pairwise methods \footnote{We only included R based implementations.}. 
 From the presented results, we note that bQCD achieves best scores in terms of SHD and SID.

 So  What exactly is the nature of the output on these undirected/directed edges, are there reported confidence scores on why the edge wasn't oriented e.g. if it were borderline or not

As such, it is reassuring that bQCD is able to correctly decide for the causal 
direction, even though other dependencies affect the pairwise (direct) causal 
relationships in such confounded dataset.
While such results are promising, a consistent hybrid method extending bQCD and quantile based methods to 
higher dimensional datasets is left for further work.
Even in this simplified setting, if one were to use \autoref{thm:causal_rule} on all pairwise adjacent vertices in a CPDAG, one may still get a PDAG (i.e., due to non-decisions).
Furthermore, extensions to higher dimensional settings could use quantile based independence tests in RESIT, or replace mean regression in CAM by quantile regression.

\begin{figure}[H]
\small
\centering
\includegraphics[width=0.65\columnwidth]{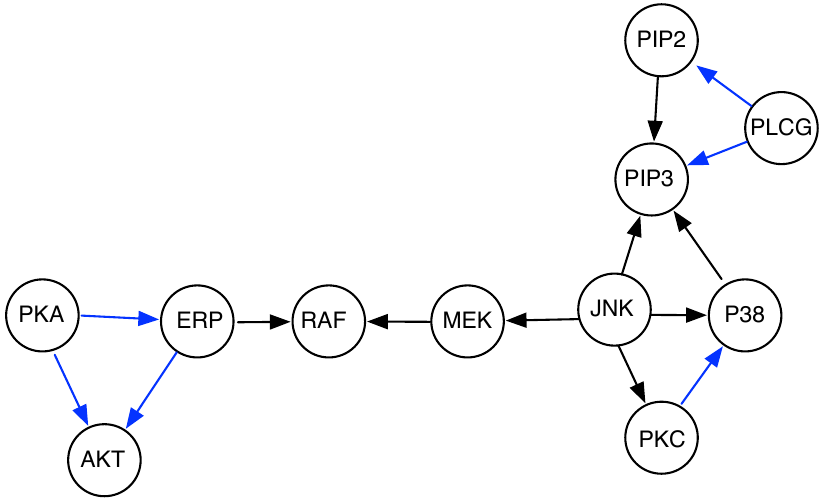}
\caption{Causal protein network as obtained by CAM's CPDAG oriented with bQCD. Blue edges are correctly oriented.}
\label{fig:sachs}
\end{figure}

\begin{table}[H]
\centering
\caption{SID and SHD for the Sachs dataset using CAM CPDAG with causal directions from pairwise methods by 10-fold cross validation. Average and standard deviations.}
\label{tab:sachs}
\resizebox{\textwidth}{!}{ 
\begin{tabular}{@{}llllllllll@{}}
\toprule
    & bQCD      & IGCI-u    & Slope     & RESIT   & CAM       & EMD       & LINGAM     & GRAN       & Random    \\ \midrule
SID & 46.5(2.5) & 50.3(1.9) & 54.5(3.3) & 51(0)   & 53.4(1.4) & 51.5(2.7) & 48.9(0.87) & 49.5(4)    & 52.5(3.6) \\
SHD & 15.1(1.6) & 15.8(1.7) & 17.6(1.2) & 14(0.3) & 15.4(1.6) & 16(1.7)   & 14.3(0.82) & 16.4(1.34) & 17.2(1.4) \\ \bottomrule
\end{tabular}
}
\end{table}

\section{Scatter plots of simulated pairs}
\label{sec:simulated_pairs}
\begin{figure}[H]
\centering
\includegraphics[width=0.55\textwidth]{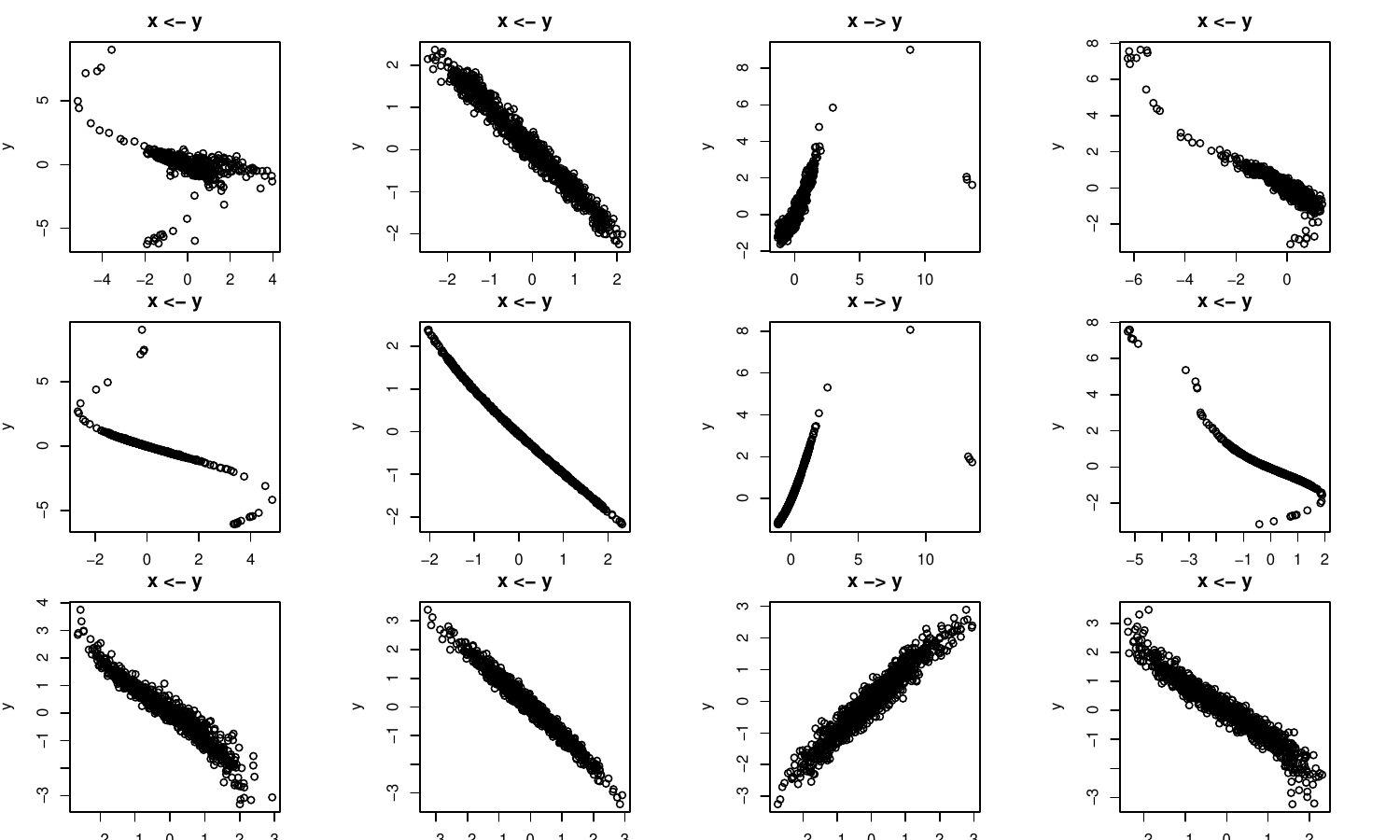}
\caption{SIM/SIM-ln/SIM-G  (first/second/third row).}
\label{fig:sim_sp}
\end{figure}

\begin{figure}
\centering
\includegraphics[width=0.55\textwidth]{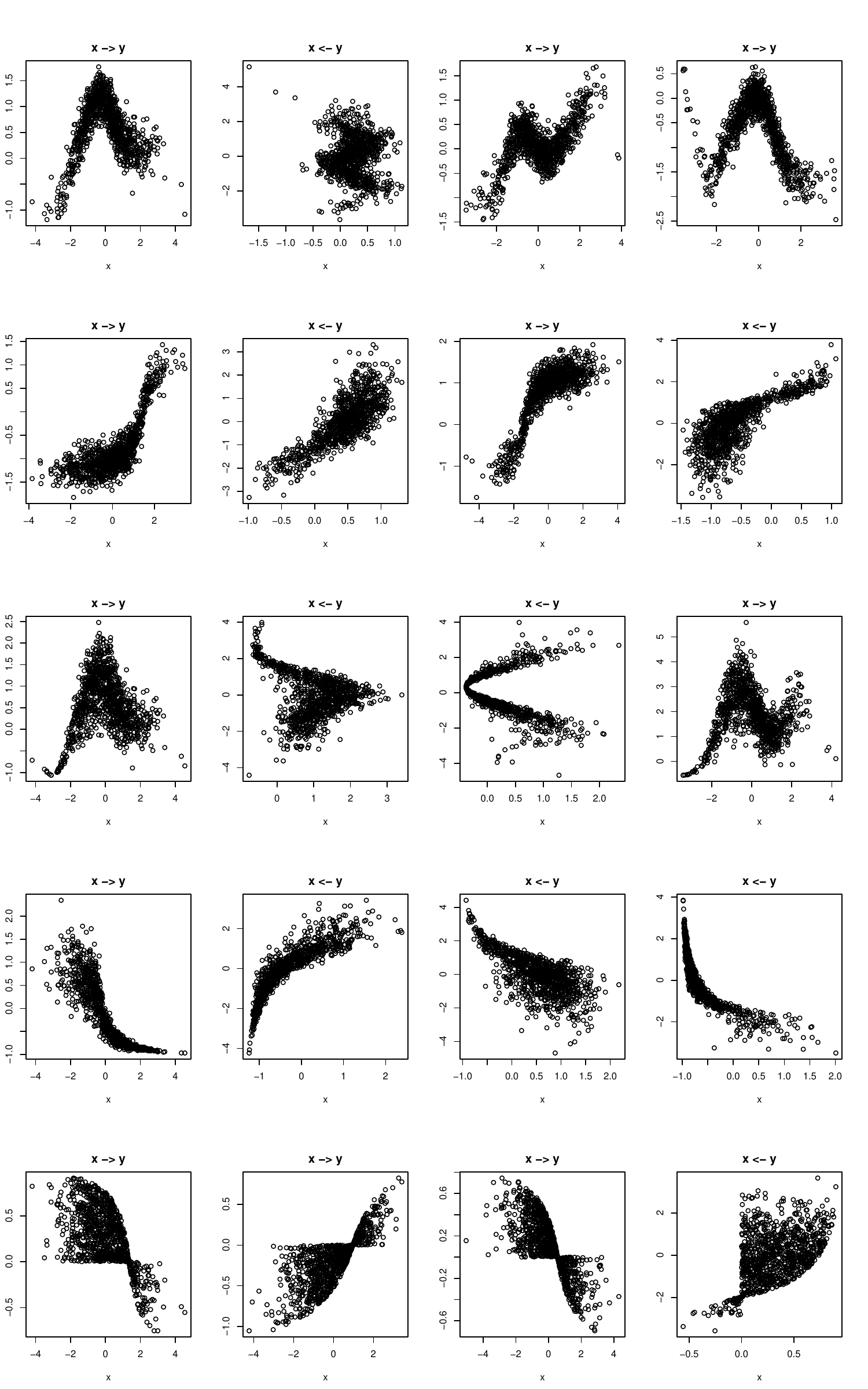}
\caption{By row AN/AN-s/LS/LS-s/MN.}
\label{fig:sim_sp2}
\end{figure}

\end{document}